\RequirePackage[T1]{fontenc}
\documentclass[journal]{IEEEtran}
\usepackage{graphicx}
\usepackage{xcolor}
\usepackage{amsmath,amssymb}
\usepackage{booktabs}
\usepackage{array}
\usepackage{adjustbox}
\usepackage{algorithm}
\usepackage{algorithmic}
\usepackage{float}
\usepackage{cite}
\usepackage[hyphens]{url}
\usepackage{xurl}
\usepackage[hidelinks]{hyperref}
="[FandolSong-Regular.otf]" at 8pt
\DeclareRobustCommand{\companyChineseName}{{\companyChinese 向上跃升（北京）科技有限公司}}
\newcommand{\companyName}{Astroup Lab}
\newcommand{\companyCity}{Beijing, China}
\newcommand{\companySite}{astroup.com}
\newcommand{\companyMailDomain}{astroup.com}
\newcommand{\extraMail}{24B903024@stu.hit.edu.cn}

\makeatletter
\def\@IEEENORMtitlevspace{0.5\baselineskip}
\def\@IEEEMINtitlevspace{0.25\baselineskip}
\makeatother

\newcommand{\qselect}{QSelect}
\newcommand{\qselectkl}{QSelect-KL}
\newcommand{\qselectta}{QSelect-TA}

\title{Compressing Onboard Language Models for Space Robots under a Behavioral-Fidelity Requirement}
\author{Jia~Deng, Zizhe~Wang, Xin~Lu, Yi~Huang, and Zehao~Jia\\[5pt]
{\small \companyName\textsuperscript{1}, \companyCity\quad
\href{https://\companySite}{\companySite}}\\[3pt]
{\small \{\href{mailto:dengjia@\companyMailDomain}{dengjia},\,
\href{mailto:wangzizhe@\companyMailDomain}{wangzizhe},\,
\href{mailto:luxin@\companyMailDomain}{luxin},\,
\href{mailto:jiazehao@\companyMailDomain}{jiazehao}\}@\companyMailDomain}\\[2pt]
{\small \href{mailto:\extraMail}{\extraMail}}\thanks{\textsuperscript{1}\companyName: \companyChineseName.}}
\hypersetup{pdftitle={Compressing Onboard Language Models for Space Robots under a Behavioral-Fidelity Requirement},pdfauthor={First Author; Second Author; Third Author}}

\begin{document}
\maketitle
\thispagestyle{IEEEtitlepagestyle}

\begin{abstract}
Language models are being explored as the interface through which space robots interpret operator commands and select among verified skills. Running them onboard calls for low-bit compression, and flight use adds requirements that generic post-training quantization leaves implicit: a fixed memory budget and storage format, a static artifact under configuration control, and behavior close to the full-precision model validated on the ground. We treat compression as a ground-segment step and study \qselect{}, which composes one static low-bit model from compatible AWQ, GPTQ, and AutoRound checkpoints by assigning each linear module to one quantizer family and scoring every substitution at the final next-token distribution. The default fidelity mode needs no labels or tuned thresholds and accepts a substitution only when the forward KL divergence from the BF16 reference decreases; an optional target-adapted mode adds answer likelihood on a small labeled set. On DeepSeek-R1-Distill-Qwen-7B, Llama-3.1-8B, and Llama-3.2-3B at 3 and 4 bits, composed models reach lower mean held-out perplexity than every single-family model in all six settings, and the fidelity mode lowers perplexity and KL further. In a matched scaffold, final-output scoring gives lower held-out KL than representation-similarity scoring in all 12 paired runs. With labels from training or development splits of the evaluated task families, the target-adapted mode raises 3-bit pooled accuracy on the two Llama models by 4.3 and 4.5 points over the best single-family model. At 3 bits, unquantized embedding and output layers make up 43--46\% of the packed artifact. All evaluations use general-domain benchmarks on a ground GPU; space-domain tasks and flight hardware are left to future work.
\end{abstract}

\section{Introduction}
\label{sec:intro}
\IEEEPARstart{S}{pace} robots are expected to work with less direct supervision. On-orbit servicing combines approach, capture, and manipulation of client spacecraft, and free-flying robots such as Astrobee serve as test beds for autonomy inside the International Space Station (ISS)~\cite{apiary2025}. Language and vision--language models offer a natural interface for such tasks: they can map an operator's request to a sequence of pre-verified skills~\cite{ahn2022} or, in end-to-end form, to robot actions~\cite{rt22023,openvla2024}. Space-specific agent frameworks now apply this interface to on-orbit servicing~\cite{spacemind2026}. These systems are usually evaluated with large models on terrestrial computers; SpaceMind, for example, uses Qwen3-VL-235B-A22B throughout~\cite{spacemind2026}, a model whose weights alone would far exceed the memory of current flight computers.

Running the model onboard removes the dependence on communication windows, round-trip delays, and limited link bandwidth, but onboard compute is scarce~\cite{furano2020}. Perseverance's primary computer, for example, is a radiation-hardened RAD750 running at up to 200~MHz with 256~MB of DRAM~\cite{nasa_perseverance_hw}; its global localization instead runs on a commercial Snapdragon~801 from the Ingenuity helicopter base station~\cite{nash2026radmit}. Language models are now reaching orbit. An AI assistant that combines ground and onboard models was delivered to China's space station in July 2025~\cite{wukong2025}, and ASTREA ran a 1.54-billion-parameter Qwen2.5 model quantized to 4 bits on a flight-qualified, CPU-only system-on-chip in an external ISS payload, with response times from about 40~s to over 8~min~\cite{mousist2025astrea}. For onboard language models, compression is therefore a prerequisite rather than an optimization.

For flight use, the relevant question is not only how small a model can become but also what the compressed model must preserve. We identify three properties. First, the artifact must fit a fixed memory budget and use a storage format that the onboard runtime supports. Second, it must be a static object that can be versioned, verified, and rolled back like other flight software. Current machine-learning guidance for space engineering does not yet cover large language models~\cite{ecssml2024}; until it does, the preparation process has to make these properties explicit. Third, its behavior must stay close to that of the full-precision reference that was validated on the ground. The third property suggests measuring compression error where the validated behavior is defined: at the model's output distribution.

Standard post-training quantization (PTQ) for large language models (LLMs) applies one quantizer family throughout a model. GPTQ uses approximate second-order error compensation~\cite{frantar2023gptq}, AWQ protects activation-salient channels by equivalent scaling~\cite{lin2024awq}, and SignRound, implemented in AutoRound, optimizes rounding and clipping~\cite{cheng2024signround}. These families embody different error models, and CALM has shown that choosing a quantization algorithm per Transformer block can help~\cite{zhang2025calm}. CALM scores candidates by linear centered kernel alignment (CKA) between intermediate representations~\cite{kornblith2019cka}. Representation similarity is a reasonable heuristic, but it is not the quantity that an acceptance test of the onboard model would examine.

\begin{figure*}[!t]
\centering
\includegraphics[width=\textwidth]{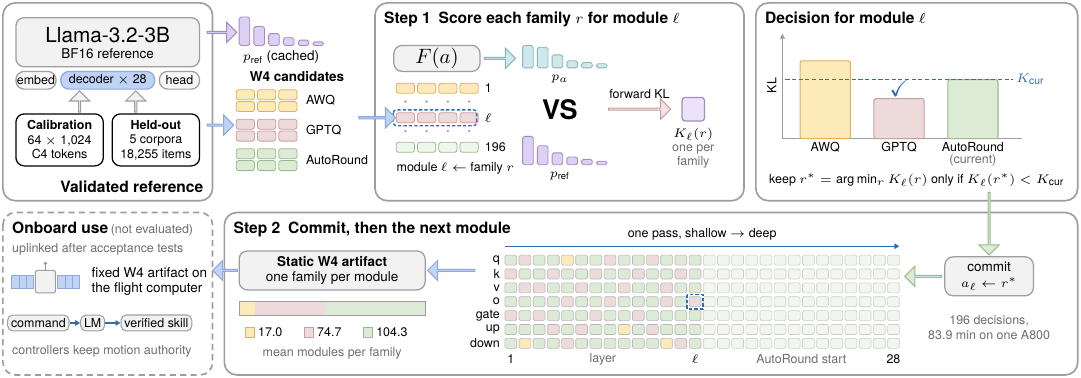}
\caption{Overview of \qselect{}, with Llama-3.2-3B at W4 as the running example. The three W4 candidates share one packed format. Step~1 gives module $\ell$ of the current model $F(a)$ to family $r$ and scores the result by the forward KL from the cached reference distribution; the lowest-KL family is kept only if it lowers the current KL (right). Step~2 commits the choice and moves on, in one shallow-to-deep pass over all 196 modules. The static artifact is uplinked after acceptance tests; onboard use is not evaluated here. Bar heights and grid colors are schematic; counts, token budgets, and times are measured values (Sections~\ref{sec:setup} and~\ref{sec:results}).}
\label{fig:pipeline}
\end{figure*}

We therefore treat compression as a ground-segment preparation step (Fig.~\ref{fig:pipeline}) and study \qselect{}, which composes one static low-bit model from compatible homogeneous checkpoints. Each linear module is assigned to one quantizer family, and every candidate substitution is scored through the final next-token distribution of the partially assigned model. In its default fidelity mode (\qselectkl{}), the procedure is label free and accepts a substitution only when the forward KL divergence from the BF16 reference strictly decreases; the rule has no tuned threshold. An optional target-adapted mode (\qselectta{}) adds a target-answer likelihood term for missions that can provide a small labeled set, with fixed margins that bound the KL increase allowed at each step. Scoring with the acceptance quantity itself, rather than a cheaper proxy, is affordable in this setting: a one-pass module-wise selection takes one to a few GPU-hours on the ground for the models studied, excluding candidate generation, is paid once per model release, and never changes the bit width or storage format of the result. The memory limits that make compression necessary also keep onboard models small, and the selection cost shrinks with the model.

Experiments on three open models from 3B to 8B parameters at 3 and 4 bits show four things. Composed models reach lower mean held-out perplexity than every single-family model in all six settings, and the fidelity mode lowers perplexity and KL further. Final-output scoring yields lower held-out KL than representation-similarity scoring in all 12 paired runs of a matched scaffold and lower KL than a local reconstruction criterion in all four tested settings. Module-wise decisions mainly lower 3-bit perplexity, at several times the ground time of layer-wise decisions, and extra passes add little. Packed artifacts are 63--70\% smaller than BF16, as for any single-family model at the same bit width, and an analytic breakdown shows that unquantized embedding and output layers make up 43--46\% of what remains at 3 bits. All experiments use general-domain language-modeling and reasoning benchmarks on a ground GPU; we do not evaluate space-domain tasks, flight hardware, or radiation effects.

Our contributions are as follows.
\begin{itemize}
\item A requirements analysis for compressing onboard language models for space robots. It states five requirements---fixed format and budget, a static artifact, behavioral fidelity to a ground-validated reference, scarce mission labels, and bounded ground compute---and maps them to design choices and to the evidence reported here (Section~\ref{sec:problem}).
\item \qselect{}, which builds on CALM's per-block algorithm selection~\cite{zhang2025calm} with a final-output score, finer decision units, a threshold-free and label-free fidelity mode, and an optional target-adapted mode (Section~\ref{sec:method}).
\item An empirical study on three open models that compares \qselect{} with single-family PTQ, representation-similarity, local-reconstruction, and whole-model selection; characterizes the trade-off between ground time and fidelity; and analyzes packed artifacts for onboard deployment (Sections~\ref{sec:results} and~\ref{sec:deployment}).
\end{itemize}

\section{Background and Related Work}
\label{sec:related}

\subsection{Language Models in Space Robotics}
Grounding language in executable skills~\cite{ahn2022} and vision--language--action models~\cite{rt22023,openvla2024} have made language a practical interface to robot behavior. In space, SpaceMind organizes an on-orbit servicing agent around natural-language skill modules and tool calls and evaluates it in simulation and on a ground robot~\cite{spacemind2026}. Space-LLaVA fine-tunes a 13B LLaVA checkpoint to adapt a vision--language model to extraterrestrial visual semantics~\cite{spacellava2025}, and language-model agents have controlled simulated spacecraft in the Kerbal Space Program Differential Games challenge~\cite{carrasco2025ksp}. These studies establish the interface; they leave open how a model of useful size can be prepared for the computers that space robots carry.

\subsection{Onboard Computing and Flight Software Assurance}
Radiation-hardened processors such as the GR740 support general flight software~\cite{gr740}, and NASA's High-Performance Spaceflight Computing processor is intended to provide at least 100 times the computational capacity of current spaceflight computers~\cite{nasa_hpsc2022}. Commercial or radiation-tolerant accelerators provide more throughput for bounded workloads, as with the Myriad~2 that ran a cloud-segmentation network on $\Phi$-sat-1~\cite{ciar2021} and the Versal XQR family~\cite{xqrversal}. Perseverance pairs its radiation-hardened primary computer with a commercial Snapdragon~801 for global localization and mitigates radiation effects in software~\cite{nash2026radmit}. ASTREA's 4-bit language model ran on the IMAGIN-e edge-computing payload~\cite{mousist2025astrea,thales_imagine2026}; at the other end of the range, a satellite carrying an NVIDIA H100 GPU has run the Gemma model in orbit~\cite{starcloud2025}. The ECSS machine-learning handbook addresses the engineering and assurance of learned components, including safety-cage architectures, but does not yet cover large language models and anticipates updates for generative AI~\cite{ecssml2024}. OPS-SAT, which ended in 2024, provided a reconfigurable payload computer for in-orbit software experiments while the bus monitored the payload and could take back control at any moment~\cite{opssat_esa}; we assume the same supervisory pattern for the onboard language model. These considerations motivate the requirements in Section~\ref{sec:problem}.

\subsection{Post-Training Quantization of LLMs}
Weight-only PTQ methods include GPTQ's second-order compensation~\cite{frantar2023gptq}, AWQ's activation-aware scaling~\cite{lin2024awq}, AdaRound's learned rounding~\cite{nagel2020adaround}, and SignRound's signed-gradient optimization~\cite{cheng2024signround}. SmoothQuant, OmniQuant, and QuaRot additionally transform or quantize activations~\cite{xiao2023smoothquant,shao2024omniquant,ashkboos2024quarot}. Recent work explores flexible low-rank approximation, spectral decomposition, and global information-flow allocation~\cite{gu2026flrq,zhao2026specquant,akbulut2026infoq}. Our setting begins after several compatible homogeneous pipelines have produced candidate checkpoints.

\subsection{Heterogeneous Assignment}
HAQ and HAWQ allocate bit widths using hardware feedback or Hessian information~\cite{wang2019haq,dong2019hawq}, and InfoQ uses a global information-flow signal for mixed-precision allocation~\cite{akbulut2026infoq}. CALM keeps the precision fixed and selects a quantization algorithm for each Transformer block with linear CKA~\cite{zhang2025calm,kornblith2019cka}. CALM and \qselect{} both assemble a model sequentially, so later decisions are conditioned on earlier ones. They differ in where a candidate is scored---intermediate block representations for CALM, the final next-token distribution for \qselect{}---and in granularity: \qselect{} supports blocks, attention or feed-forward components, and individual linear modules. Our CALM baseline is an independent matched-pool reimplementation, not an official reproduction.

\subsection{Output-Aware Calibration and Modular Composition}
Knowledge distillation transfers output-distribution information~\cite{hinton2015distilling}, and output-adaptive calibration uses model-output distortion to guide PTQ~\cite{edalati2025oac}. AdaRound and BRECQ instead optimize weight-, layer-, or block-local reconstruction losses~\cite{nagel2020adaround,li2021brecq}. \qselect{} uses the final-output KL only for discrete assignment, not for training candidates. The construction is also related to model stitching~\cite{bansal2021stitching}, although we combine compatible modules from quantized versions of the same checkpoint without learning a stitching layer.

\section{Ground-to-Orbit Model Preparation}
\label{sec:problem}

\subsection{Deployment Workflow}
Fig.~\ref{fig:pipeline} summarizes the setting. On the ground, a full-precision (BF16) reference model is validated for its mission role, for example interpreting operator commands and selecting among verified manipulation or inspection skills. Several PTQ pipelines that share a storage format then produce homogeneous low-bit checkpoints. \qselect{} composes one static model from their modules; the result is packed, recorded with its configuration, and compared with the reference by acceptance tests before it is preloaded before launch or uplinked as a software update. Onboard, the model runs at a supervisory rate inside the autonomy stack. Its outputs pass through the same checks and fallback logic as any other planner output, and conventional controllers keep authority over motion.

\begin{table*}[!t]
\centering
\footnotesize
\caption{Deployment requirements for an onboard language model and how they are addressed}
\label{tab:requirements}
\begin{tabular}{@{}p{0.27\textwidth}p{0.42\textwidth}p{0.24\textwidth}@{}}
\toprule
Requirement & Design choice in \qselect{} & Evidence in this paper \\
\midrule
R1 Fixed memory budget and a storage format the onboard runtime supports & All candidates share bit width, group size, and packed integer format; selection never changes the bit width & Table~\ref{tab:artifact}, Fig.~\ref{fig:artifact} \\
R2 Static artifact under configuration control & One candidate per module, fixed after selection; no runtime selector, routing, or ensemble; the assignment vector and checkpoint hashes identify the artifact & By construction (Section~\ref{sec:export}) \\
R3 Behavior close to the ground-validated reference & Candidates are scored by the forward KL from the reference next-token distribution on calibration text; acceptance can check the same quantity on held-out text & Tables~\ref{tab:main_results}--\ref{tab:cheap_criteria} \\
R4 Few or no mission labels & Fidelity mode needs only unlabeled text; the target-adapted mode uses a small labeled set & Tables~\ref{tab:main_results} and~\ref{tab:cheap_criteria} \\
R5 Bounded, one-time ground compute & Decision granularity and pass count trade ground time for fidelity & Fig.~\ref{fig:cost}, Table~\ref{tab:granularity} \\
\bottomrule
\end{tabular}
\end{table*}

\subsection{Requirements}
Table~\ref{tab:requirements} lists five requirements and the corresponding design choices.

\emph{R1: fixed memory budget and supported format.} The model must fit the memory left after flight software and buffers, and the onboard runtime must be able to execute its storage format. All candidates therefore share the bit width, group size, and symmetric packed integer format, so the footprint of the quantized layers is fixed before selection. AWQ modules additionally carry a static per-channel input scale. The composition introduces no new numerical format, although it may use family-specific module wrappers (Section~\ref{sec:export}). Keeping one bit width also keeps one integer kernel and one memory layout to verify, whereas mixed-precision assignment~\cite{wang2019haq,dong2019hawq} multiplies both; this is why we select among quantizer families rather than among bit widths.

\emph{R2: static artifact under configuration control.} Each module keeps exactly one selected candidate. There is no runtime selector, routing, or ensemble, so the onboard network is fixed and can be characterized once on the ground. The assignment vector, with 196 or 224 entries for the models studied here, together with the candidate checkpoint and manifest hashes, identifies the artifact for versioning and rollback.

\emph{R3: behavioral fidelity to the validated reference.} Ground validation establishes how the reference behaves on mission-relevant inputs. The compressed model should depart from that behavior as little as possible. We measure the departure by the forward KL divergence from the reference next-token distribution to the compressed model's distribution, averaged over text positions; it is zero only when the two distributions agree at every evaluated position. Forward KL weights each position by the reference's own probabilities, so it penalizes most the loss of outputs that the reference considers likely. At one position, the total-variation distance between the two distributions is the smallest probability, over couplings, that the two models sample different next tokens; by Pinsker's and Jensen's inequalities, an average KL of $K$ nats bounds its mean over positions by $\sqrt{K/2}$. The bound holds on reference text rather than on the model's own generations, so it motivates the score rather than guaranteeing free-running behavior. An acceptance test can apply the same quantity to held-out or mission text, alongside task tests, while selection optimizes it on separate calibration text; the remaining gap between selection and acceptance therefore lies in the data rather than in the criterion. For a validated system, a change that raises task accuracy but also raises KL is an unvalidated change in behavior, which is why answer supervision is optional (R4). This paper sets no numerical KL threshold; a mission would derive one from its own validation campaign.

\emph{R4: scarce mission labels.} Labeled mission data are expensive and often unavailable when a model is prepared. The fidelity mode therefore uses only unlabeled text and the reference model. When a small labeled set exists, the target-adapted mode can use it, but it then departs from the reference on purpose, and acceptance tests must cover that departure.

\emph{R5: bounded, one-time ground compute.} Selection runs on ground GPUs once per model release, whereas onboard resources are consumed throughout the mission. Spending GPU-hours on the ground is acceptable if it buys fidelity, but the cost should be predictable and adjustable. Decision granularity and the number of passes provide that control, and the cost shrinks with the model, which onboard memory keeps small (Section~\ref{sec:groundtime}).

\subsection{Problem Statement}
Let $F_{\mathrm{ref}}$ be the validated BF16 reference and let $H^{(1)},\ldots,H^{(R)}$ be homogeneous low-bit checkpoints with the same bit width and format. The task is to choose one candidate family for every quantized linear module so that the composed model minimizes the fidelity score $K$ of Section~\ref{sec:score}, using only unlabeled text. Optionally, a bounded increase in $K$ may be traded for a lower target-answer loss $T$. Because every candidate for a module has the same bit width and group size, the memory budget is fixed by construction. Exhaustive search is impractical: module-wise assignment of Llama-3.1-8B over three families has $3^{224}$ possibilities. \qselect{} therefore uses a greedy search whose cost grows linearly with the number of decision units and candidates.

\section{\qselect{}}
\label{sec:method}
\qselect{} operates on a precomputed bank of matched-format module candidates. It scores temporary substitutions in the current partially assigned model, commits accepted updates according to a fixed search protocol, and exports the resulting static assignment.

\subsection{Candidate Bank and Assignment Space}
Let $r\in\{1,\ldots,R\}$ index a quantizer family, $s$ a calibration-bank replicate, and $l\in\{1,\ldots,L\}$ a target linear module. For every $r$ and $s$ we first construct a homogeneous quantized checkpoint $H_s^{(r)}$. Its module $l$ supplies a family-conditioned candidate
\begin{equation}
\mathcal B_{l,s}^{(r)}=\widehat W_{l,s}^{(r)}.
\label{eq:candidate_bank}
\end{equation}
During assignment, $\widehat W_{l,s}^{(r)}$ is stored as a dense effective fake-quantized weight. The notation retains the calibration-bank replicate and the homogeneous pipeline: a candidate is not assumed to be an independently generated function of $W_l$ alone. For AWQ, the selection-time effective weight is
\begin{equation}
\widehat W_{l,s}^{(\mathrm{AWQ})}
=Q_b\!\left(W_l\operatorname{diag}(d_{l,s})\right)
\operatorname{diag}(d_{l,s})^{-1},
\label{eq:awq_fold}
\end{equation}
where $d_{l,s}$ is the input-channel scale and $Q_b$ is the common symmetric group-wise weight quantizer. A family is supported only when its transformed parameters and static state can be encapsulated as a locally swappable unit with a compatible input--output interface; global rotations, cross-layer transformations, activation quantization, and incompatible activation formats are outside this scope.

Let $\mathcal C_l\subseteq\{1,\ldots,R\}$ index the candidates available for module $l$. An assignment $a=(a_1,\ldots,a_L)$ chooses $a_l\in\mathcal C_l$ for every module. Let $\mathcal G$ partition the modules into decision units. The legal families for unit $G\in\mathcal G$ are
\begin{equation}
\mathcal C_G=\bigcap_{l\in G}\mathcal C_l,
\end{equation}
and the granularity-constrained assignment space is
\begin{equation}
\mathcal A_{\mathcal G}=\left\{a:\;a_l\in\mathcal C_l,\;a_l=a_{l'}\ \text{for all }l,l'\in G,\ G\in\mathcal G\right\}.
\end{equation}
Singleton units give module-wise assignment; one unit per decoder block gives layer-wise assignment; grouping the query, key, value, and output projections separately from the gate, up, and down projections gives component-wise assignment.

\subsection{Final-Output Fidelity Score}
\label{sec:score}
For an assignment $a$, let $F_s(a)$ be the resulting quantized model. On non-padding next-token positions $\Omega_{\mathrm{KL}}$ we compute the forward KL divergence from the reference distribution to the quantized-model distribution:
\begin{equation}
K_s(a)=\frac{1}{|\Omega_{\mathrm{KL}}|}
\sum_{(x,t)\in\Omega_{\mathrm{KL}}}
D_{\mathrm{KL}}\!\left(
 p_{\mathrm{ref},t}(\cdot\mid x)\,\|\,
 p_{a,t}(\cdot\mid x)
\right).
\label{eq:kl_score}
\end{equation}
The distributions use the full vocabulary and temperature one. Reference logits are cached on the CPU in FP16, transferred to FP32, and normalized with \texttt{log\_softmax}; model logits and KL accumulation use FP32. The cache requires $O(|\Omega_{\mathrm{KL}}|V)$ host memory for vocabulary size $V$. The reference is evaluated once; every later evaluation runs only the candidate model.

\subsection{Fidelity Mode}
The fidelity mode, \qselectkl{}, is the default. At each decision unit it evaluates every legal family, proposes the one with the lowest $K_s$, and accepts it whenever its score is strictly lower than that of the current assignment (Algorithm~\ref{alg:fidelity}). No labels, margins, or tuned thresholds are involved, so the search is a greedy block-coordinate descent on Eq.~\eqref{eq:kl_score}: the score of the committed assignment never increases. Exact ties follow the fixed order AWQ, GPTQ, AutoRound.

\begin{algorithm}[t]
\caption{\qselectkl{}: one-pass fidelity mode}
\label{alg:fidelity}
\begin{algorithmic}[1]
\REQUIRE Candidate bank $\{\mathcal B_{l,s}^{(r)}\}$, decision units $\mathcal G$ in a fixed order, initial assignment $a$.
\STATE $K_{\mathrm{cur}}\leftarrow K_s(a)$
\FOR{decision unit $G$ in the fixed order}
  \STATE $K_G(r)\leftarrow K_s(a[G\leftarrow r])$ for every $r\in\mathcal C_G$
  \STATE $r^*\leftarrow\arg\min_{r\in\mathcal C_G}K_G(r)$
  \IF{$K_G(r^*)<K_{\mathrm{cur}}$}
    \STATE $a\leftarrow a[G\leftarrow r^*]$; $K_{\mathrm{cur}}\leftarrow K_G(r^*)$
  \ENDIF
\ENDFOR
\RETURN static assignment $a$ and model $F_s(a)$.
\end{algorithmic}
\end{algorithm}

\subsection{Target-Adapted Mode}
Some missions can supply prompts $x$ with target answers $y=(y_1,\ldots,y_{|y|})$, for example procedure questions with known answers. The target-adapted mode, \qselectta{}, uses the token-averaged target-answer negative log-likelihood (NLL)
\begin{equation}
T(a)=-\frac{
\sum_{(x,y)\in\mathcal D_{\mathrm{proxy}}}
\sum_{j=1}^{|y|}\log p_a(y_j\mid x,y_{<j})
}{\sum_{(x,y)\in\mathcal D_{\mathrm{proxy}}}|y|}.
\label{eq:proxy_nll}
\end{equation}
This is a teacher-forced answer-likelihood signal, not free-running generation or accuracy. At each unit, \qselectta{} first forms a near-best-KL shortlist,
\begin{equation}
\mathcal S_G=\left\{r\in\mathcal C_G:
K_G(r)\leq1.01\min_{r'\in\mathcal C_G}K_G(r')\right\},
\label{eq:shortlist}
\end{equation}
and proposes the shortlist candidate with the lowest $T$. The proposal is accepted through either a fidelity-improvement branch,
\begin{equation}
K_G(r^*)\leq0.999K_{\mathrm{cur}},
\label{eq:accept_fidelity}
\end{equation}
or a target-adaptation branch,
\begin{equation}
K_G(r^*)\leq1.002K_{\mathrm{cur}}
\quad\land\quad
T_G(r^*)\leq0.997T_{\mathrm{cur}}.
\label{eq:accept_task}
\end{equation}
The 1\% shortlist, 0.1\% KL-improvement margin, 0.2\% KL tolerance, and 0.3\% NLL-improvement margin were fixed before the evaluated comparisons and shared across all settings; they were not derived theoretically or exhaustively swept. The margins bound how far a single accepted step can move away from the reference. The full procedure, including lazy NLL recomputation, is given in Algorithm~\ref{alg:qselect} in the Appendix.

\subsection{Search Protocol}
The main protocol starts from the homogeneous AutoRound assignment, scans decoder blocks from shallow to deep, and processes the query, key, value, output, gate, up, and down projections in that order. It performs one pass. Candidate evaluation is functional, so temporary insertions do not alter the committed assignment. Initialization, granularity, traversal order, and pass budget are protocol variables, and Section~\ref{sec:robustness} reports their effects.

\subsection{Ground Cost and Static Export}
\label{sec:export}
Let $U=|\mathcal G|$, let $R_u$ and $S_u$ be the candidate and shortlist sizes for unit $u$, and let $C_{\mathrm{model}}(N)$ be the cost of a full-model pass over $N$ scored tokens. Assignment requires approximately
\begin{equation}
O\!\left(\sum_{u=1}^{U}R_u C_{\mathrm{model}}(N_{\mathrm{KL}})
+\sum_{u=1}^{U}S_u C_{\mathrm{model}}(N_{\mathrm{proxy}})\right),
\end{equation}
where the second term is absent in the fidelity mode. Candidate generation and $O(RP)$ candidate-bank storage are separate costs, where $P$ is the number of stored target-module parameters. With fixed width, no prefix or suffix caching, $U$ proportional to depth $D$, and a full forward pass proportional to $D$, the direct module-wise implementation grows approximately quadratically with depth. All of this cost is incurred on the ground.

Native export packs the selected modules after assignment and does not alter the selected families. Each packed module stores low-bit integer codes, group scales, and the zero-point field of the packed layout, which is constant under the symmetric format; AWQ modules additionally retain $d_{l,s}$ as static metadata and apply $x\mapsto x\operatorname{diag}(d_{l,s})^{-1}$ before the packed linear operator. The exported model is therefore statically composed of family-specific modules, with no dynamic selector or ensemble. The assignment vector, the candidate checkpoint hashes, and the calibration manifest hashes form the configuration record of the artifact.

\section{Experimental Setup}
\label{sec:setup}
\emph{Scope.} The experiments evaluate the ground segment of Fig.~\ref{fig:pipeline}. Our evaluation contains no space-domain data; general-domain language-modeling and reasoning benchmarks stand in for the reference behavior that a mission would validate. Each job runs on one NVIDIA A800-SXM4-80GB GPU.

\emph{Models and low-bit format.} We use \href{https://huggingface.co/deepseek-ai/DeepSeek-R1-Distill-Qwen-7B}{\nolinkurl{deepseek-ai/DeepSeek-R1-Distill-Qwen-7B}}~\cite{guo2025deepseekr1,deepseek2025modelcard} (DS-R1-Qwen-7B), \href{https://huggingface.co/meta-llama/Meta-Llama-3.1-8B}{\nolinkurl{meta-llama/Meta-Llama-3.1-8B}}~\cite{dubey2024llama3}, and \href{https://huggingface.co/meta-llama/Llama-3.2-3B}{\nolinkurl{meta-llama/Llama-3.2-3B}}~\cite{meta2024llama32}. BF16 references are loaded without low-bit PTQ. W3A16 and W4A16 (W3 and W4) denote symmetric 3- or 4-bit weights, BF16 activations, no zero point, and input-channel groups of 128. We quantize the query, key, value, output, gate, up, and down projections; embeddings, normalization layers, and the output head remain BF16. Unless stated otherwise, results use dense effective fake-quantized weights; packed artifacts are examined in Section~\ref{sec:deployment}.

\emph{Candidate generation.} The three families use the same replicate-specific C4 manifest~\cite{raffel2020exploring} of 128 packed sequences of 2,048 tokens. They do not use matched compute or identical effective token budgets. GPTQ uses GPTQModel 7.1.0 with symmetric quantization, \texttt{desc\_act=false}, \texttt{true\_sequential=true}, and damping 0.05. AutoRound 0.13.0 uses the SignRound recipe of that release with 200 iterations, batch size one, gradient accumulation one, and \texttt{packed\_tensors}; unlisted options use that release's defaults. Our matched-format AWQ implementation gathers at most 1,024 activation tokens per module, searches a 20-point scaling grid, and applies Eq.~\eqref{eq:awq_fold}. These configurations are candidate generators in a shared output format, not claims about each method's strongest published configuration.

\emph{Calibration banks and selector data.} Replicates 17, 23, and 41 index candidate generation and the 64-sequence, 1,024-token C4 sample used for KL scoring. The labeled proxy and held-out evaluation manifests are fixed. For DS-R1-Qwen-7B, the proxy uses 128 GSM8K training, 128 MATH training, and 256 MMLU development examples; the Llama proxy additionally includes 128 ARC-Challenge training and 256 HellaSwag training examples. GPQA is never included. A shared-core control applies the DS proxy to Llama-3.1-8B.

\emph{CALM reimplementation.} CALM-reimpl.\ follows sequential block-wise assembly and selects one family per block with biased linear CKA. It compares the BF16 and candidate outputs of \texttt{mlp.down\_proj}, the pre-residual feed-forward contribution, over 64 C4 sequences of 1,024 tokens, accumulating global centered feature statistics in 1,024-token chunks. The candidate pool is AWQ/GPTQ/AutoRound rather than CALM's original AWQ/GPTQ/SmoothQuant pool.

\emph{Intrinsic evaluation.} Each replicate is evaluated on fixed manifests of 128 sequences of length 2,048 from C4, WikiText-2~\cite{merity2016pointer}, and three local corpora denoted Code, Math, and Wiki. Tables report the arithmetic average of the five corpus-level perplexity (PPL) values and the corresponding maximum, together with corpus-averaged and maximum KL. We label these columns \emph{Avg.} and \emph{Max-corpus} because they are descriptive summaries rather than pooled language-modeling metrics.

\emph{Downstream evaluation.} The suite contains GPQA Diamond (198)~\cite{rein2023gpqa}, ARC-Challenge (1,172)~\cite{clark2018arc}, GSM8K (1,319)~\cite{cobbe2021gsm8k}, HellaSwag (1,024)~\cite{zellers2019hellaswag}, MATH-500 (500)~\cite{lightman2024verify}, and MMLU (14,042)~\cite{hendrycks2021mmlu}. The protocol is zero-shot, with no chat template or system prompt. Multiple-choice tasks are reformatted as letter-answer questions and scored by average answer-letter log probability, so these scores are not directly comparable with standard continuation-likelihood harness values. GSM8K and MATH-500 use greedy generation with at most 256 new tokens and normalized final-answer matching. Micro pools 18,255 examples and is dominated by MMLU; Macro is the unweighted mean of the six task accuracies. The Appendix gives further protocol details.

\emph{Modes evaluated.} \qselectkl{} is the default mode, but most protocol studies were run with \qselectta{}: the granularity, traversal-order, initialization, candidate-pool, pass-budget, and whole-model controls, the packed export, and the module-wise 7B/8B selection times all use the target-adapted mode. \qselectkl{} was run with the main protocol in all six settings. For the four 7B/8B settings, its results are available as paired changes relative to \qselectta{} (Table~\ref{tab:cheap_criteria}), and its absolute values in Table~\ref{tab:main_results} are derived from them ($^\dagger$, without standard deviations); on Llama-3.2-3B they are measured directly. Because the fidelity mode evaluates a subset of the target-adapted mode's forward passes (Section~\ref{sec:export}), the target-adapted selection times bound those of the fidelity mode under the same protocol.

\section{Results}
\label{sec:results}

\begin{table*}[!t]
\centering
\scriptsize
\setlength{\tabcolsep}{2.45pt}
\caption{Main intrinsic and downstream results (mean $\pm$ sample standard deviation over calibration-bank replicates 17, 23, and 41)}
\label{tab:main_results}
\begin{adjustbox}{max width=\textwidth}
\begin{tabular}{lllrrrrrr}
\toprule
Model & Bits & Method & Avg. PPL$\downarrow$ & Max-corpus PPL$\downarrow$ & Avg. KL$\downarrow$ & Max-corpus KL$\downarrow$ & Micro$\uparrow$ & Macro$\uparrow$ \\
\midrule
DS-R1-Qwen-7B & BF16 & Reference & 26.68 & 44.32 & 0.000 & 0.000 & 49.64 & 41.19 \\
\cmidrule(lr){2-9}
 & W3 & AWQ & 51.13$\pm$1.98 & 88.63$\pm$3.69 & 0.885$\pm$0.016 & 1.058$\pm$0.021 & 34.70$\pm$0.90 & 29.31$\pm$0.40 \\
 & W3 & GPTQ & 33.14$\pm$1.06 & 56.31$\pm$2.47 & 0.436$\pm$0.005 & \textbf{0.498}$\pm$0.016 & 37.97$\pm$0.99 & 31.22$\pm$0.18 \\
 & W3 & AutoRound & 35.03$\pm$0.69 & 58.94$\pm$1.13 & 0.474$\pm$0.006 & 0.562$\pm$0.005 & \textbf{41.85}$\pm$0.14 & 33.25$\pm$0.48 \\
 & W3 & \qselectta{} & \textbf{30.70}$\pm$0.26 & \textbf{51.15}$\pm$0.61 & \textbf{0.420}$\pm$0.004 & \textbf{0.498}$\pm$0.010 & 41.77$\pm$1.95 & \textbf{34.02}$\pm$2.04 \\
 & W3 & \qselectkl{} & 29.23$^\dagger$ & -- & 0.389$^\dagger$ & -- & 41.62$^\dagger$ & 34.77$^\dagger$ \\
\cmidrule(lr){2-9}
 & W4 & AWQ & 27.68$\pm$0.26 & 46.16$\pm$0.33 & 0.132$\pm$0.002 & 0.160$\pm$0.004 & 46.94$\pm$0.52 & 39.82$\pm$0.47 \\
 & W4 & GPTQ & 28.09$\pm$0.15 & 47.27$\pm$0.43 & \textbf{0.092}$\pm$0.001 & \textbf{0.106}$\pm$0.003 & 48.60$\pm$0.13 & 40.81$\pm$0.34 \\
 & W4 & AutoRound & 29.29$\pm$0.47 & 49.33$\pm$0.84 & 0.124$\pm$0.003 & 0.151$\pm$0.004 & 48.81$\pm$0.07 & 40.67$\pm$0.21 \\
 & W4 & \qselectta{} & \textbf{27.33}$\pm$0.50 & \textbf{45.49}$\pm$0.88 & 0.1059$\pm$0.0016 & 0.1250$\pm$0.0019 & \textbf{48.83}$\pm$0.12 & \textbf{41.19}$\pm$0.20 \\
 & W4 & \qselectkl{} & 26.87$^\dagger$ & -- & 0.0913$^\dagger$ & -- & 48.85$^\dagger$ & 40.74$^\dagger$ \\
\midrule
Llama-3.1-8B & BF16 & Reference & 7.01 & 10.05 & 0.000 & 0.000 & 59.68 & 45.44 \\
\cmidrule(lr){2-9}
 & W3 & AWQ & 14.97$\pm$0.54 & 22.08$\pm$0.88 & 0.734$\pm$0.029 & 0.898$\pm$0.034 & 26.69$\pm$0.53 & 19.64$\pm$0.23 \\
 & W3 & GPTQ & 10.39$\pm$0.13 & 14.71$\pm$0.16 & 0.432$\pm$0.014 & 0.550$\pm$0.007 & 41.36$\pm$2.13 & 30.23$\pm$1.23 \\
 & W3 & AutoRound & 11.01$\pm$0.17 & 15.96$\pm$0.18 & 0.432$\pm$0.013 & 0.521$\pm$0.026 & 43.54$\pm$0.89 & 31.24$\pm$0.76 \\
 & W3 & \qselectta{} & \textbf{9.79}$\pm$0.05 & \textbf{14.03}$\pm$0.02 & \textbf{0.3486}$\pm$0.0075 & \textbf{0.4231}$\pm$0.0116 & \textbf{47.81}$\pm$0.53 & \textbf{36.24}$\pm$0.70 \\
 & W3 & \qselectkl{} & 9.58$^\dagger$ & -- & 0.3336$^\dagger$ & -- & 46.91$^\dagger$ & 35.04$^\dagger$ \\
\cmidrule(lr){2-9}
 & W4 & AWQ & 7.74$\pm$0.05 & 11.08$\pm$0.09 & 0.110$\pm$0.006 & 0.127$\pm$0.006 & 56.42$\pm$0.35 & 42.98$\pm$0.95 \\
 & W4 & GPTQ & 7.689$\pm$0.005 & 11.188$\pm$0.011 & 0.0937$\pm$0.0011 & 0.1204$\pm$0.0002 & 56.63$\pm$0.19 & 43.56$\pm$0.29 \\
 & W4 & AutoRound & 7.94$\pm$0.05 & 11.54$\pm$0.03 & 0.108$\pm$0.005 & 0.131$\pm$0.014 & 56.38$\pm$0.27 & \textbf{44.01}$\pm$0.58 \\
 & W4 & \qselectta{} & \textbf{7.60}$\pm$0.02 & \textbf{10.93}$\pm$0.02 & \textbf{0.0835}$\pm$0.0011 & \textbf{0.1061}$\pm$0.0026 & \textbf{57.34}$\pm$0.14 & 43.73$\pm$0.23 \\
 & W4 & \qselectkl{} & 7.55$^\dagger$ & -- & 0.0810$^\dagger$ & -- & 57.04$^\dagger$ & 43.33$^\dagger$ \\
\midrule
Llama-3.2-3B & BF16 & Reference & 8.70 & 12.32 & 0.000 & 0.000 & 51.03 & 38.06 \\
\cmidrule(lr){2-9}
 & W3 & AWQ & 24.44$\pm$2.56 & 43.71$\pm$5.63 & 0.962$\pm$0.088 & 1.305$\pm$0.123 & 25.82$\pm$0.58 & 19.81$\pm$0.26 \\
 & W3 & GPTQ & 16.66$\pm$1.93 & 26.14$\pm$4.84 & 0.668$\pm$0.124 & 0.799$\pm$0.166 & 33.71$\pm$1.23 & 26.30$\pm$0.98 \\
 & W3 & AutoRound & 15.13$\pm$0.33 & 23.08$\pm$0.60 & 0.527$\pm$0.015 & 0.675$\pm$0.019 & 36.13$\pm$0.89 & 25.84$\pm$0.88 \\
 & W3 & \qselectta{} & 12.92$\pm$0.90 & 19.82$\pm$2.02 & 0.408$\pm$0.058 & 0.512$\pm$0.095 & \textbf{40.60}$\pm$0.36 & \textbf{30.10}$\pm$0.67 \\
 & W3 & \qselectkl{} & \textbf{12.21}$\pm$0.09 & \textbf{18.20}$\pm$0.06 & \textbf{0.367}$\pm$0.011 & \textbf{0.434}$\pm$0.008 & 39.02$\pm$0.49 & 29.06$\pm$0.42 \\
\cmidrule(lr){2-9}
 & W4 & AWQ & 9.76$\pm$0.04 & 14.15$\pm$0.06 & 0.118$\pm$0.005 & 0.152$\pm$0.006 & 48.46$\pm$0.20 & \textbf{37.28}$\pm$0.54 \\
 & W4 & GPTQ & 9.73$\pm$0.16 & 13.71$\pm$0.32 & 0.111$\pm$0.010 & 0.143$\pm$0.023 & 48.24$\pm$0.15 & 35.12$\pm$0.65 \\
 & W4 & AutoRound & 9.92$\pm$0.03 & 13.99$\pm$0.07 & 0.118$\pm$0.003 & 0.155$\pm$0.009 & 47.71$\pm$0.50 & 35.10$\pm$0.80 \\
 & W4 & \qselectta{} & 9.46$\pm$0.05 & 13.77$\pm$0.31 & 0.092$\pm$0.006 & 0.125$\pm$0.025 & \textbf{48.66}$\pm$0.80 & 36.07$\pm$0.67 \\
 & W4 & \qselectkl{} & \textbf{9.37}$\pm$0.10 & \textbf{13.56}$\pm$0.45 & \textbf{0.088}$\pm$0.007 & \textbf{0.117}$\pm$0.028 & 48.50$\pm$0.69 & 36.30$\pm$0.76 \\
\bottomrule
\end{tabular}
\end{adjustbox}
\par\smallskip
\begin{minipage}{\textwidth}\scriptsize
Reference rows are unquantized BF16 models evaluated once. Avg.\ PPL and Avg.\ KL are arithmetic means over five corpora within each replicate; Max-corpus PPL and KL are the corresponding maxima. Micro pools all 18,255 task examples; Macro averages the six task accuracies. Bold marks the best measured mean within each setting and does not indicate statistical significance. $^\dagger$Derived mean: the \qselectta{} mean shifted by the paired change of \qselectkl{} in Table~\ref{tab:cheap_criteria} (relative for PPL and KL, in points for accuracy). Standard deviations and maxima are unavailable, the last digit can differ by one from a direct computation, and derived values are not bolded, although several derived PPL and KL means are lower than the bold ones. For Llama-3.2-3B, where both are available, derived and measured means agree to within 0.01.
\end{minipage}
\end{table*}

\subsection{Composed Versus Single-Family Models}
Table~\ref{tab:main_results} compares both modes with the homogeneous candidate generators. At W3, \qselectta{} lowers mean Avg. PPL by 7.4\%, 5.8\%, and 14.6\% relative to the best single-family model on DS-R1-Qwen-7B, Llama-3.1-8B, and Llama-3.2-3B, and it raises pooled (Micro) accuracy on the two Llama models by 4.3 and 4.5 points over the best single family. These task gains use labeled proxy examples from training or development splits of five of the six evaluated Llama task families (Section~\ref{sec:setup}), so they measure adaptation to known tasks rather than label-free fidelity. On DS-R1-Qwen-7B, PPL and KL improve in every calibration replicate, but the Micro differences from AutoRound are $+1.11$, $+1.12$, and $-2.47$ points (Table~\ref{tab:dsw3_paired} in the Appendix), so the task result depends on the calibration bank. Changes at W4 are smaller: DS W4 improves PPL but not KL, and both Llama models show metric-dependent task trade-offs.

The fidelity mode lowers mean Avg. PPL and Avg. KL relative to \qselectta{} in all six settings (Table~\ref{tab:cheap_criteria}). Because \qselectta{} is already below every single-family Avg. PPL mean, \qselectkl{} has the lowest mean Avg. PPL among the configurations of Table~\ref{tab:main_results} in every setting, and the lowest Avg. KL in five; in DS W4 its derived KL (0.0913) is comparable to that of homogeneous GPTQ (0.092). The task cost of removing answer supervision is concentrated at Llama-3.2 W3, where every paired replicate loses Micro accuracy and the means fall by 1.57 Micro and 1.03 Macro points; at W4 the changes are $-0.16$ Micro and $+0.23$ Macro. Even without labels, the fidelity mode exceeds the best single-family Micro accuracy at W3 on both Llama models, by 2.9 points on Llama-3.2-3B and 3.4 points (derived) on Llama-3.1-8B.

\subsection{Why Score the Final Output?}
\label{sec:criteria}
Three comparisons isolate the choice of scoring signal. Held-out KL partly measures what a final-output selector optimizes, so we also report perplexity on the five held-out corpora and task accuracy, which the selectors in the first two comparisons do not optimize.

\emph{Representation similarity.} Table~\ref{tab:matched_compact} replaces CKA with final-distribution KL inside the same sequential layer-wise scaffold, with the same candidate bank and token budget. The KL selector in this control is a guarded layer-wise variant that accepts a change only when it lowers KL by at least 0.1\% (Eq.~\eqref{eq:accept_fidelity}); it is not the threshold-free fidelity mode. The final-output score lowers held-out Avg. KL in each of the twelve paired calibration-bank runs (Table~\ref{tab:matched_seed_deltas} in the Appendix). PPL and task accuracy also improve on DS W3 and on both Llama settings, whereas DS W4 trades lower KL for slightly higher PPL and lower task means. Because selection and evaluation both use KL, this control supports a fidelity-oriented scoring choice within the scaffold rather than a universal superiority claim. The final-output score is not much more expensive at equal granularity: layer-wise selection takes 19.4--21.6 minutes with KL and 16.9--18.1 minutes with CKA (Table~\ref{tab:matched_scoring}).

\begin{table}[t]
\centering
\scriptsize
\setlength{\tabcolsep}{3.2pt}
\caption{Fidelity mode and a local reconstruction criterion, both relative to the module-wise \qselectta{} mean. PPL and KL are relative changes of three-replicate means; accuracy values are percentage points. Negative PPL/KL and positive accuracy favor the listed configuration. Local-MSE was run on the 7B/8B settings only.}
\label{tab:cheap_criteria}
\begin{tabular}{llrrrr}
\toprule
Setting & Configuration & $\Delta$PPL & $\Delta$KL & $\Delta$Micro & $\Delta$Macro \\
\midrule
DS W3 & \qselectkl{} & -4.8\% & -7.5\% & -0.15 & +0.75 \\
 & Local-MSE & +8.0\% & +1.7\% & -2.16 & -1.20 \\
DS W4 & \qselectkl{} & -1.7\% & -13.8\% & +0.02 & -0.45 \\
 & Local-MSE & +2.9\% & -13.5\% & +0.04 & +0.32 \\
Llama-3.1 W3 & \qselectkl{} & -2.1\% & -4.3\% & -0.90 & -1.20 \\
 & Local-MSE & -0.7\% & +4.3\% & -3.60 & -3.30 \\
Llama-3.1 W4 & \qselectkl{} & -0.6\% & -3.0\% & -0.30 & -0.40 \\
 & Local-MSE & -0.9\% & +5.8\% & -0.80 & -0.60 \\
Llama-3.2 W3 & \qselectkl{} & -5.5\% & -10.1\% & -1.57 & -1.03 \\
Llama-3.2 W4 & \qselectkl{} & -1.0\% & -4.5\% & -0.16 & +0.23 \\
\bottomrule
\end{tabular}
\end{table}

\begin{table}[t]
\centering
\scriptsize
\caption{CKA versus guarded final-distribution KL (0.1\% margin) inside the same sequential layer-wise scaffold. Deltas are final-KL minus CALM-reimpl.; PPL/KL are relative percentages and accuracy values are percentage points.}
\label{tab:matched_compact}
\begin{tabular}{lrrrr}
\toprule
Setting & $\Delta$PPL & $\Delta$KL & $\Delta$Micro & $\Delta$Macro \\
\midrule
DS W3 & -1.42\% & -9.96\% & +0.74 & +1.17 \\
DS W4 & +2.03\% & -5.07\% & -0.03 & -0.46 \\
Llama-3.1 W3 & -6.26\% & -19.14\% & +5.45 & +4.29 \\
Llama-3.1 W4 & -0.81\% & -17.14\% & +0.68 & +0.88 \\
\bottomrule
\end{tabular}
\end{table}

\emph{Local reconstruction.} The Local-MSE selector collects 8,192 stratified input tokens per module from the BF16 reference and unconditionally chooses the family with the lowest mean-squared error between reference and candidate module outputs. It uses static reference activations, so it is not a contextual search and has no initialization or acceptance threshold. It takes about 0.8 minutes on the DS settings. Table~\ref{tab:cheap_criteria} shows that it ends with higher Avg. KL than the fidelity mode in all four tested settings (marginally on DS W4), higher Avg. PPL in three, and lower Micro accuracy in three. Local-MSE is thus a useful quick reference, not a substitute for the output criterion.

\emph{Whole-model selection.} Applying the same KL shortlist, answer NLL, and acceptance rule to whole homogeneous models selects GPTQ in all four 7B/8B settings, AutoRound in all Llama-3.2 W3 replicates, and GPTQ or AutoRound in two or one Llama-3.2 W4 replicates (Table~\ref{tab:homogeneous_control}). In the five settings where all replicates agree, module-wise \qselectta{} lowers Avg. PPL by 1.22--14.62\% and raises Micro and Macro by 0.23--6.45 and 0.17--6.01 points relative to the selected family. This control does not match the search capacity or path dependence of module-wise selection, but it shows that choosing one family with the same rule does not recover the module-wise results.

\begin{table}[t]
\centering
\scriptsize
\caption{Whole-model family choices made by the same KL-shortlist and answer-NLL rule. Counts are over calibration-bank replicates 17, 23, and 41.}
\label{tab:homogeneous_control}
\begin{tabular}{lrrr}
\toprule
Setting & AWQ & GPTQ & AutoRound \\
\midrule
DS W3 & 0 & 3 & 0 \\
DS W4 & 0 & 3 & 0 \\
Llama-3.1 W3 & 0 & 3 & 0 \\
Llama-3.1 W4 & 0 & 3 & 0 \\
Llama-3.2 W3 & 0 & 0 & 3 \\
Llama-3.2 W4 & 0 & 2 & 1 \\
\bottomrule
\end{tabular}
\end{table}

\subsection{Ground Time Versus Fidelity}
\label{sec:groundtime}
\begin{figure*}[!t]
\centering
\includegraphics[width=\textwidth]{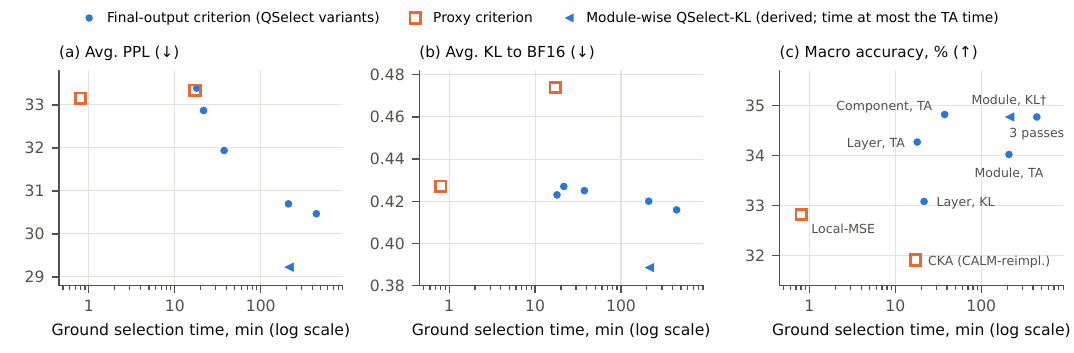}
\caption{Ground selection time versus fidelity and task accuracy on DS-R1-Qwen-7B W3; labels in (c) apply to all panels. Filled circles use the final output (layer-, component-, and module-wise \qselectta{}, the guarded layer-wise KL selector, and module-wise \qselectta{} with three passes); hollow squares are proxy criteria (Local-MSE and CALM-reimpl.\ with CKA). All values come from Tables~\ref{tab:main_results}, \ref{tab:cheap_criteria}, \ref{tab:granularity}, \ref{tab:matched_scoring}, and~\ref{tab:search_controls}. Local-MSE and three-pass metrics are derived from reported relative changes. Times for Local-MSE and the layer, component, and module scans are the approximate values reported for the DS settings (about 0.8, 18, 35--40, and 210 minutes); the bar on the component point spans 35--40 minutes. The triangle is module-wise \qselectkl{}, derived from Table~\ref{tab:cheap_criteria}; its time is not reported, so it is drawn at the module-wise \qselectta{} time, which bounds it from above.}
\label{fig:cost}
\end{figure*}

Fig.~\ref{fig:cost} places the evaluated selectors on one time axis for DS W3, and Table~\ref{tab:granularity} gives the granularity study. DS layer, component, and module scans take roughly 18, 35--40, and 210 minutes. Finer units are not uniformly better: modules give the lowest W3 PPL, components the highest DS task means, and at W4 layers give the lowest DS KL and the lowest Llama-3.1 PPL. At layer granularity, \qselectta{} gives higher Micro accuracy than CALM-reimpl.\ in all four settings and lower Avg. KL in three, with DS W4 essentially tied (0.0987 versus 0.0986); CALM-reimpl.\ gives slightly lower Avg. PPL on both DS settings. Three passes triple the candidate evaluations and raise selection time to 7.2--9.5 hours, changing Avg. PPL by only $-0.12$\% to $-0.75$\% (Table~\ref{tab:search_controls}); one pass remains the default. On one A800 GPU, a one-pass module-wise \qselectta{} run for Llama-3.1-8B W3 takes 3.12 hours, and \qselectkl{} on Llama-3.2-3B averages 83.9 minutes, both excluding candidate generation, which is not reported here. Exact final-output scoring is therefore far slower than a local proxy such as Local-MSE, which reuses static reference activations, but at equal granularity it adds only a few minutes over representation-similarity scoring, which also runs the assembled model (Section~\ref{sec:criteria}); most of the module-wise time comes from the number of decisions. The search itself is deliberately simple: a greedy coordinate descent with one forward pass per candidate. Its absolute cost stays modest because each evaluation is a forward pass of the model being prepared, and onboard memory keeps that model small: Llama-3.2-3B, the smallest model studied here, is still about twice the size of the 1.54-billion-parameter model flown on ASTREA~\cite{mousist2025astrea}. The budget that makes compression necessary thus also keeps exact scoring affordable. The cost stays on the ground, because the exported model has the bit width and format of a single-family model apart from the AWQ input scales, but it recurs with every model update, because a new calibration bank changes the artifact (Section~\ref{sec:deployment}-C).

\begin{table*}[!t]
\centering
\scriptsize
\setlength{\tabcolsep}{3.1pt}
\caption{CALM comparison and \qselect{} decision granularity. CALM-reimpl. is an independently implemented layer-wise CKA selector with the matched candidate pool. The three \qselectta{} rows keep the candidate bank, AutoRound initialization, scores, thresholds, forward order, one-pass budget, and calibration-bank replicates fixed while changing the decision unit. Values are mean $\pm$ sample standard deviation; bold marks the best displayed mean within each setting and is descriptive.}
\label{tab:granularity}
\begin{adjustbox}{max width=\textwidth}
\begin{tabular}{lllrrrr}
\toprule
Setting & Selector & Unit & Avg. PPL$\downarrow$ & Avg. KL$\downarrow$ & Micro$\uparrow$ & Macro$\uparrow$ \\
\midrule
DS-R1-Qwen-7B W3 & CALM-reimpl. & Layer & 33.34$\pm$1.81 & 0.474$\pm$0.015 & 38.49$\pm$1.05 & 31.91$\pm$0.75 \\
 & \qselectta{} & Layer & 33.39$\pm$0.65 & 0.423$\pm$0.006 & 41.97$\pm$0.66 & 34.27$\pm$0.84 \\
 & \qselectta{} & Component & 31.94$\pm$0.95 & 0.425$\pm$0.008 & \textbf{42.50}$\pm$0.35 & \textbf{34.82}$\pm$1.22 \\
 & \qselectta{} & Module & \textbf{30.70}$\pm$0.26 & \textbf{0.420}$\pm$0.004 & 41.77$\pm$1.95 & 34.02$\pm$2.04 \\
\midrule
DS-R1-Qwen-7B W4 & CALM-reimpl. & Layer & 27.41$\pm$0.23 & 0.0986$\pm$0.0014 & 48.55$\pm$0.41 & 41.07$\pm$0.73 \\
 & \qselectta{} & Layer & 27.77$\pm$0.62 & 0.0987$\pm$0.0013 & 48.81$\pm$0.11 & 41.08$\pm$0.18 \\
 & \qselectta{} & Component & 27.84$\pm$0.30 & 0.103$\pm$0.001 & \textbf{49.09}$\pm$0.09 & \textbf{41.27}$\pm$0.17 \\
 & \qselectta{} & Module & \textbf{27.33}$\pm$0.50 & 0.106$\pm$0.002 & 48.83$\pm$0.12 & 41.19$\pm$0.20 \\
\midrule
Llama-3.1-8B W3 & CALM-reimpl. & Layer & 10.56$\pm$0.01 & 0.451$\pm$0.008 & 39.28$\pm$3.25 & 28.84$\pm$2.18 \\
 & \qselectta{} & Layer & 9.94$\pm$0.08 & 0.364$\pm$0.005 & 45.61$\pm$0.50 & 33.78$\pm$0.60 \\
 & \qselectta{} & Component & 9.85$\pm$0.13 & 0.353$\pm$0.008 & 46.40$\pm$1.20 & 34.39$\pm$1.48 \\
 & \qselectta{} & Module & \textbf{9.79}$\pm$0.05 & \textbf{0.349}$\pm$0.008 & \textbf{47.81}$\pm$0.53 & \textbf{36.24}$\pm$0.70 \\
\midrule
Llama-3.1-8B W4 & CALM-reimpl. & Layer & 7.60$\pm$0.09 & 0.099$\pm$0.007 & 56.30$\pm$0.28 & 42.91$\pm$0.47 \\
 & \qselectta{} & Layer & \textbf{7.55}$\pm$0.04 & 0.0842$\pm$0.0007 & 56.63$\pm$0.77 & 42.87$\pm$0.91 \\
 & \qselectta{} & Component & 7.62$\pm$0.05 & 0.0837$\pm$0.0017 & 57.25$\pm$0.29 & 43.20$\pm$0.63 \\
 & \qselectta{} & Module & 7.60$\pm$0.02 & \textbf{0.0835}$\pm$0.0011 & \textbf{57.34}$\pm$0.14 & \textbf{43.73}$\pm$0.23 \\
\bottomrule
\end{tabular}
\end{adjustbox}
\end{table*}

\subsection{Robustness of the Ground Procedure}
\label{sec:robustness}
Traversal order has a smaller aggregate effect than initialization. Random and reverse scans change Avg. PPL by at most 0.10 in three settings; on DS W3 the increases are 0.77 and 0.53, and mean Macro changes by at most 0.67 points across all orders (Table~\ref{tab:order_sensitivity}). Changing the restricted-pool Llama-3.1 W3 start from GPTQ to AWQ, by contrast, increases Avg. PPL by 1.18 and lowers Micro and Macro by 5.49 and 5.14 points (Table~\ref{tab:pool_init}). Adding AutoRound to the pool has the clearest value at W3: across both models and both AWQ and GPTQ starts, it lowers Avg. PPL by 0.32--1.24 and raises Micro by 2.37--4.02 points, while W4 changes are smaller and mixed (Tables~\ref{tab:initialization} and~\ref{tab:pool_init}).

The selected models remain heterogeneous. On Llama-3.2-3B, the mean AWQ/GPTQ/AutoRound counts of \qselectkl{} are 6.0/65.0/125.0 at W3 and 17.0/74.7/104.3 at W4. Across the three DS W3 calibration banks, pairwise assignment agreement is 73.5--76.0\%, and 124 of 196 modules agree in all three. Selection therefore depends on the candidate bank as well as on the model, a point we return to in Section~\ref{sec:deployment}.

\section{Deployment Analysis for Space Robots}
\label{sec:deployment}
This section interprets the measured results for onboard use. It adds an analytic storage model but no new experiments.

\subsection{What Occupies the Packed Artifact}
Table~\ref{tab:artifact} reports the packed exports of the calibration-replicate-17 assignments. Artifact reduction compares the serialized artifact directory with the BF16 reference directory. Packing changes held-out metrics only slightly relative to the fake-quantized model: $|\Delta\mathrm{PPL}|\leq0.277\%$ and $|\Delta\mathrm{KL}|\leq1.314\%$, with drift defined as $\Delta_z=100\,[z(M_{\mathrm{nat}})-z(M_{\mathrm{fq}})]/z(M_{\mathrm{fq}})$ for the packed model $M_{\mathrm{nat}}$ and the fake-quantized model $M_{\mathrm{fq}}$.

\begin{figure}[t]
\centering
\includegraphics[width=\columnwidth]{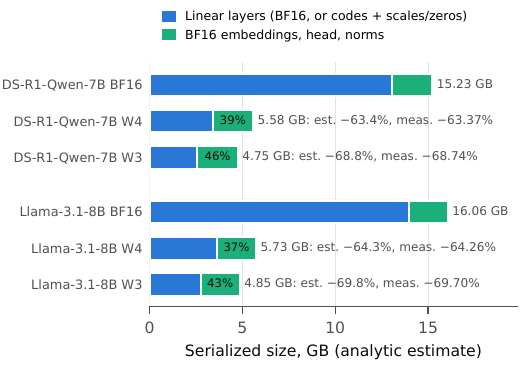}
\caption{Analytic composition of the serialized artifact. Green segments are the BF16 embedding, output head, and normalization layers; the percentages give their share of the quantized artifact. Each label compares the estimated reduction with the measured one in Table~\ref{tab:artifact}.}
\label{fig:artifact}
\end{figure}

To see where the remaining bytes are, we model the artifact from the public model configurations. Quantized linear layers store $b$-bit codes, a packed $b$-bit zero point and an FP16 scale for every group of 128 weights, and an int32 group index per input channel; embeddings, the output head, normalization weights, and biases stay in BF16. This model reproduces the four measured reductions to within 0.08 percentage points (Table~\ref{tab:artifact}). It also shows that the BF16 embedding and output layers, which the experiments never quantize, make up 43--46\% of the 3-bit artifact and 37--39\% of the 4-bit artifact (Fig.~\ref{fig:artifact}). Further reductions for an onboard memory budget must therefore address those layers, for example through embedding and head quantization or vocabulary restriction; we have not evaluated either.

\begin{table}[t]
\centering
\scriptsize
\setlength{\tabcolsep}{2.6pt}
\caption{Packed export of the calibration-replicate-17 assignments (\qselectta{}). Artifact and peak-CUDA entries are reductions relative to the BF16 reference; $\Delta$PPL and $\Delta$KL are packed-versus-fake-quantized drift. $^\dagger$From the analytic storage model (Section~\ref{sec:deployment}-A), not measured.}
\label{tab:artifact}
\begin{tabular}{lrrrrrr}
\toprule
 & \multicolumn{3}{c}{Artifact reduction} & Peak & \multicolumn{2}{c}{Drift} \\
\cmidrule(lr){2-4}\cmidrule(lr){6-7}
Setting & Measured & Est.$^\dagger$ & BF16 share$^\dagger$ & CUDA & $\Delta$PPL & $\Delta$KL \\
\midrule
DS W3 & 68.74\% & 68.79\% & 45.9\% & 26.3\% & -0.029\% & -0.089\% \\
DS W4 & 63.37\% & 63.40\% & 39.1\% & 24.2\% & -0.028\% & +0.024\% \\
Llama-3.1 W3 & 69.70\% & 69.78\% & 43.3\% & 27.9\% & +0.033\% & -1.314\% \\
Llama-3.1 W4 & 64.26\% & 64.30\% & 36.7\% & 25.9\% & -0.277\% & -0.232\% \\
\bottomrule
\end{tabular}
\end{table}

\subsection{Runtime Memory}
Peak CUDA memory falls by only 24.2--27.9\% (Table~\ref{tab:artifact}), although the serialized artifact shrinks by roughly a factor of three. The absolute sizes also matter: by the analytic model, the 7B/8B artifacts still occupy about 4.8~GB at 3 bits and 5.6--5.7~GB at 4 bits. They are far beyond the 256~MB of a RAD750-class computer, call for accelerator-class payload computers, and are several times larger than the 1.54-billion-parameter model flown on ASTREA. The measurement uses batch one, sequence length 2,048, and no key--value cache, includes model loading, and runs the packed forward path used here, which dequantizes weights during execution and is not an optimized serving kernel. A flight deployment would need kernels that compute directly from packed weights and a memory plan for activations and the key--value cache, measured on flight-representative hardware. Latency matters as much as memory there: ASTREA's 1.54-billion-parameter 4-bit model needed from about 40~s to over 8~min per response on a CPU-only flight system-on-chip~\cite{mousist2025astrea}. We leave these measurements to future work.

\subsection{Choosing an Operating Point}
\emph{Bit width.} At W4, composed models stay close to the reference: \qselectta{} is within 0.81, 2.34, and 2.37 Micro points of BF16 on DS-R1-Qwen-7B, Llama-3.1-8B, and Llama-3.2-3B. At W3 the gaps are 7.87, 11.87, and 10.43 points even after composition. For command interpretation, where the validated reference behavior is the specification, W4 is the realistic default; W3 fits only when memory is the binding constraint and task-level acceptance tests can quantify the loss. At W4, composition buys fidelity rather than accuracy: relative to the best single family, the fidelity mode lowers Avg. KL from 0.0937 to 0.0810$^\dagger$ on Llama-3.1-8B and from 0.111 to 0.088 on Llama-3.2-3B and matches it on DS-R1-Qwen-7B (0.0913$^\dagger$ versus 0.092), while its Micro accuracy stays within half a point of the best single-family value.

\emph{Mode.} The fidelity mode is the natural default for R3: it needs no labels and, among the configurations of Table~\ref{tab:main_results}, gives the lowest Avg. PPL in every setting and the lowest Avg. KL in five of six. The target-adapted mode is appropriate when the mission supplies labeled examples and its acceptance tests cover the resulting departure from the reference. At Llama-3.2 W3, for instance, the fidelity mode lowers Avg. KL by 10.1\% relative to \qselectta{} but loses 1.57 Micro points.

\emph{Updates.} Because candidates are calibration-bank specific, re-running selection with a new bank changes the artifact substantially: across the DS W3 banks, only 124 of 196 module choices agree in all three. Even modules whose family is unchanged receive new weights, because each bank regenerates the candidates, so an update replaces the whole quantized artifact. A model update should therefore be treated as a new artifact with full acceptance testing rather than as a patch, and the configuration record (Section~\ref{sec:export}) makes the previous artifact recoverable for rollback.

\subsection{What Flight Use Would Add}
Three questions remain open. First, the reference behavior here is general-domain text and reasoning; a mission would validate the reference on its own command corpus, procedures, and skill library, and the fidelity score should then be computed on that text. Second, latency, energy, and thermal load must be measured on flight-representative processors. Third, packed formats change how a radiation-induced upset propagates: a corrupted group scale or zero point affects all 128 weights of its group, so these metadata may deserve stronger protection than the codes. A corrupted AWQ input scale affects a whole input channel of its module, so the exposure also depends on the assignment. None of these questions is answered by the present experiments.

\section{Limitations}
\label{sec:limitations}
\emph{No space-domain evaluation.} All results use general-domain benchmarks on a ground GPU. We do not evaluate space-robot commands, procedures, or skill libraries, flight hardware, latency, energy, or radiation effects.

\emph{Scope of the method.} The evidence covers fixed-bit symmetric weight-only candidates with group size 128 for 3B--8B text-only models. Candidates inherit homogeneous calibration paths and are dense fake-quantized weights during assignment; family token budgets and compute differ. Global rotations, cross-layer transforms, activation quantization, incompatible runtime formats, vision--language models, and models larger than 8B are not demonstrated. Both Llama references are base models; a command interface would start from an instruction-tuned reference.

\emph{Metrics.} The arithmetic five-corpus PPL average and raw maximum are descriptive rather than pooled metrics. The downstream protocol is custom, \qselectta{} uses labels from the evaluated task families and token-weights answer NLL differently from Micro and Macro accuracy, GPQA is the only proxy-unseen Llama task, and three calibration-bank replicates do not measure proxy or test-sampling uncertainty. Some \qselectkl{} values in Table~\ref{tab:main_results} are derived from paired changes rather than measured directly.

\emph{Search.} Assignments depend on initialization, order, passes, the target-adapted margins, and the candidate bank. One pass is a cost choice, not convergence, and rejected NLL proposals do not trigger runner-up tests. Candidate generation and storage are extra costs, and native export is executable but not an optimized serving kernel. The argument that exact scoring is affordable rests on the 3B--8B sizes studied; larger models or frequent re-preparation would call for prefix caching or a cheaper screening step before final-output scoring, for example a local Fisher-information approximation of the same KL, which we do not evaluate.

\section{Conclusion}
\label{sec:conclusion}
Onboard language models for space robots need more than a small footprint: the compressed model must be a static artifact in a supported format whose behavior stays close to the reference validated on the ground. Treating compression as a ground-segment step, \qselect{} composes one low-bit model from compatible quantizer families and scores every substitution through the final next-token distribution. Its label-free fidelity mode has no tuned threshold and reaches lower mean held-out perplexity than every single-family model and the target-adapted mode in all six settings; final-output scoring beats representation-similarity and local-reconstruction criteria on held-out KL; and ground time can be traded against fidelity through decision granularity. The packed artifacts are 63--70\% smaller than BF16, and nearly half of what remains at 3 bits (43--46\%) is the unquantized embedding and output layers rather than the composed modules. Evaluating the fidelity criterion on mission text and measuring the composed models on flight-representative hardware are the next steps.

\section*{Acknowledgment}
The authors thank the developers of the open models and quantization libraries used in this work.

\IEEEtriggeratref{40}

\clearpage
\appendices

\section{Target-Adapted Algorithm}
\label{app:ta}
Algorithm~\ref{alg:qselect} gives the one-pass \qselectta{} procedure. The implementation caches the current KL and NLL scores. If a proposal is accepted without evaluating NLL, the NLL cache is invalidated and recomputed lazily before its next use. If the minimum-NLL proposal fails both acceptance branches, no runner-up is tried.

\begin{algorithm}[!t]
\caption{One-pass \qselectta{} assignment}
\label{alg:qselect}
\begin{algorithmic}[1]
\REQUIRE Candidate bank $\{\mathcal B_{l,s}^{(r)}\}$, units $\mathcal G$, feasible initial assignment $a$, fixed order, and Eqs.~\eqref{eq:shortlist}--\eqref{eq:accept_task}.
\STATE $K_{\mathrm{cur}}\leftarrow K_s(a)$; $T_{\mathrm{cur}}\leftarrow T(a)$; $T_{\mathrm{valid}}\leftarrow\mathrm{true}$.
\FOR{decision unit $G$ in the fixed order}
  \STATE Let $r_{\mathrm{cur}}$ be the family currently assigned to $G$.
  \FOR{$r\in\mathcal C_G$}
    \STATE $K_G(r)\leftarrow K_s(a[G\leftarrow r])$.
  \ENDFOR
  \STATE Form $\mathcal S_G$ with Eq.~\eqref{eq:shortlist}.
  \IF{$|\mathcal S_G|=1$}
    \STATE Let $r^*$ be its sole family; mark $T_G(r^*)$ unevaluated.
  \ELSE
    \IF{$\neg T_{\mathrm{valid}}$}
      \STATE $T_{\mathrm{cur}}\leftarrow T(a)$; $T_{\mathrm{valid}}\leftarrow\mathrm{true}$.
    \ENDIF
    \FOR{$r\in\mathcal S_G$}
      \STATE $T_G(r)\leftarrow T(a[G\leftarrow r])$, reusing $T_{\mathrm{cur}}$ for $r_{\mathrm{cur}}$.
    \ENDFOR
    \STATE Let $r^*$ minimize $T_G(r)$ under fixed tie breaking.
  \ENDIF
  \IF{$r^*\neq r_{\mathrm{cur}}$ and Eq.~\eqref{eq:accept_fidelity} holds}
    \STATE $a\leftarrow a[G\leftarrow r^*]$; $K_{\mathrm{cur}}\leftarrow K_G(r^*)$.
    \IF{$T_G(r^*)$ was evaluated}
      \STATE $T_{\mathrm{cur}}\leftarrow T_G(r^*)$; $T_{\mathrm{valid}}\leftarrow\mathrm{true}$.
    \ELSE
      \STATE $T_{\mathrm{valid}}\leftarrow\mathrm{false}$.
    \ENDIF
  \ELSIF{$r^*\neq r_{\mathrm{cur}}$, $T_G(r^*)$ was evaluated, and Eq.~\eqref{eq:accept_task} holds}
    \STATE $a\leftarrow a[G\leftarrow r^*]$; $K_{\mathrm{cur}}\leftarrow K_G(r^*)$; $T_{\mathrm{cur}}\leftarrow T_G(r^*)$.
    \STATE $T_{\mathrm{valid}}\leftarrow\mathrm{true}$.
  \ENDIF
\ENDFOR
\RETURN $a$ and $F_s(a)$.
\end{algorithmic}
\end{algorithm}

\section{Protocol Details}
\label{app:protocol}
The calibration-bank replicates are 17, 23, and 41. Each replicate determines the candidate-generation C4 manifest and the separate C4 sample used for KL scoring; the labeled proxy and held-out evaluation manifests are fixed. Candidate generation uses 128 sequences of length 2,048. KL scoring uses 64 sequences of length 1,024. The five-corpus suite uses 128 sequences per corpus at length 2,048 and evaluation seed 1009. Manifest hashes and candidate checkpoint hashes are recorded in the run metadata.

The proxy builder takes fixed dataset rows without per-run resampling. GSM8K uses the first 128 training rows and the string following ``\#\#\#\#''; MMLU uses the first 256 development rows and answer letters; ARC-Challenge and HellaSwag use the first 128 and 256 training rows, respectively, with the original option order; MATH cycles through seven training subjects until 128 parseable final answers are collected. Multiple-choice proxy prompts end in \texttt{Answer:}, and the target continuation is a leading space followed by the gold letter. GSM8K uses \texttt{Question: ... Answer:} and MATH uses \texttt{Problem: ... Final answer:}, each followed only by the canonical final-answer string. No reasoning trace is included.

Downstream evaluation is zero-shot with raw text prompts. Multiple-choice prompts list lettered options and end in \texttt{Answer:}; the score is the average continuation log probability of a leading-space answer letter. HellaSwag is reformatted as a multiple-choice question with activity and context fields. GSM8K uses \texttt{Question: ... Answer: Let's think step by step.}, and MATH-500 uses \texttt{Problem: ... Solution: Let's solve it step by step. The final answer is}. Generation is greedy with at most 256 new tokens and stops at the tokenizer EOS token. GSM8K uses the final numeric string; MATH-500 first extracts the last boxed expression and then applies whitespace, comma, dollar-sign, and case normalization with a last-number fallback. This custom protocol supports matched internal comparison but should not be read as a standard lm-evaluation-harness score.

The local Code corpus is built from source files in the evaluation repository (Python, C/C++, CUDA, Rust, Go, Java, JavaScript, and TypeScript), excluding common dependency and cache directories. The Math builder uses MMLU-Pro question text~\cite{wang2024mmlupro} when available, and the Wiki builder uses WikiText-103 training text. The frozen files support internally repeatable comparison, but the resolved upstream revisions, license metadata, file counts, language composition, and complete deduplication provenance were not preserved, so these corpora should be treated as local robustness sets rather than standard public benchmarks.

Software versions are PyTorch 2.10.0+cu128, Transformers 5.8.0, Datasets 4.0.0, Accelerate 1.13.0, GPTQModel 7.1.0, AutoRound 0.13.0, and Safetensors 0.7.0. Each reported job uses one NVIDIA A800-SXM4-80GB GPU. The launch metadata records checkpoint names but does not pin model and tokenizer commit revisions, which limits reproducibility.

\section{Additional Results}
\label{app:results}

\subsection{Per-Task Results}
Table~\ref{tab:per_task_results} lists selected per-task accuracies. The selected rows expose task-level trade-offs but do not compare all candidate families; for Llama, the proxy-unseen group contains only GPQA.

\subsection{Matched Layer-Wise Scoring}
Table~\ref{tab:matched_scoring} reports the full-budget CALM-reimpl.\ and final-distribution KL variants discussed in Section~\ref{sec:criteria}; Table~\ref{tab:matched_seed_deltas} gives the paired Avg.-KL differences. Selection time excludes candidate generation. The guarded layer-wise KL selector accepts a change only under the 0.1\% margin of Eq.~\eqref{eq:accept_fidelity}.

\subsection{Objective Ablations}
Module-wise \qselectkl{} selects the minimum-KL candidate and accepts any strict KL decrease. NLL-only ranks all candidates by target-answer NLL and accepts a replacement when NLL improves by at least 0.3\%, without a KL condition. NLL-first + KL guard ranks all candidates by NLL, removes the KL shortlist, and applies the default acceptance conditions in Eqs.~\eqref{eq:accept_fidelity}--\eqref{eq:accept_task}. All variants retain the same replicate-specific candidate bank, initialization, order, and proxy data (Table~\ref{tab:objective_ablation}). NLL-only improves both task aggregates only on DS W3 and degrades PPL and KL in all four evaluated 7B/8B settings, so answer supervision shifts the operating point rather than uniformly improving it.

\subsection{Paired DS-R1-Qwen-7B W3 Analysis}
In each calibration-bank replicate, the task baseline with the highest Micro and Macro value is AutoRound. For PPL and KL, the baseline is selected separately within each replicate and metric. Negative intrinsic deltas and positive task deltas favor \qselectta{} (Table~\ref{tab:dsw3_paired}). The replicate-41 Micro decrease is driven primarily by a 2.91-point MMLU decrease relative to AutoRound; because MMLU contributes 14,042 of 18,255 examples, this term accounts for approximately 2.23 points of the total Micro change. Pairwise assignment agreement is 73.5--76.0\%, and 124/196 modules agree across all three replicates. Among the 22 modules for which replicates 17 and 23 agree but replicate 41 differs, 12 change from AutoRound to GPTQ. The observed association does not determine which replacements cause the task decrease.

\subsection{Search Budget and Proxy Control}
Table~\ref{tab:search_controls} reports differences from the corresponding one-pass main result. Every three-pass row uses exactly three full scans. Shared-core uses the DS proxy subset for Llama-3.1-8B.

\subsection{Initialization Sensitivity}
Table~\ref{tab:initialization} varies only the homogeneous starting assignment. Different starts retain different family majorities, and no start is uniformly best across fidelity and task metrics.

\subsection{Candidate Pool $\times$ Initialization}
Together with the full-pool AWQ and GPTQ starts in Table~\ref{tab:initialization}, Table~\ref{tab:pool_init} completes a $2\times2$ comparison of candidate pool and initialization. The restricted pool contains AWQ and GPTQ only; all other protocol choices remain fixed. The large Llama-3.1 W3 difference between AWQ and GPTQ starts shows that a restricted-pool result cannot be interpreted as a pure source-removal effect.

\subsection{Traversal-Order Sensitivity}
Table~\ref{tab:order_sensitivity} compares the fixed forward protocol with one fixed random permutation (order seed 2027) and the exact reverse of the full module list. Each row is paired with the same calibration-bank replicate of the forward main result. Deltas are variant minus forward, so negative PPL/KL and positive accuracy favor the alternative order.

\begin{table*}[p]
\centering
\scriptsize
{\setlength{\tabcolsep}{2.05pt}
\caption{Selected per-task downstream accuracy (\%). The 7B/8B rows show the AutoRound initialization baseline and \qselectta{}; the 3B rows show all three homogeneous candidates plus \qselectkl{} and \qselectta{}. This table is not exhaustive for the 7B/8B settings. Micro pools all examples, whereas Macro averages tasks.}
\label{tab:per_task_results}
\begin{adjustbox}{max width=\textwidth}
\begin{tabular}{lllrrrrrrrr}
\toprule
Model & Bits & Method & GPQA & ARC-C & GSM8K & HellaSwag & MATH500 & MMLU & Micro & Macro \\
\midrule
DS-R1-Qwen-7B & BF16 & Reference & 31.31 & 68.94 & 23.81 & 50.59 & 20.80 & 51.67 & 49.64 & 41.19 \\
\cmidrule(lr){2-11}
 & W3 & AutoRound & 23.74$\pm$2.20 & 59.41$\pm$0.62 & 21.13$\pm$2.27 & 34.34$\pm$0.46 & 16.87$\pm$2.42 & 44.03$\pm$0.37 & 41.85$\pm$0.14 & 33.25$\pm$0.48 \\
 & W3 & \qselectta{} & 23.23$\pm$1.34 & 58.62$\pm$3.08 & 22.52$\pm$2.82 & 36.20$\pm$4.27 & 19.93$\pm$0.90 & 43.62$\pm$1.85 & 41.77$\pm$1.95 & 34.02$\pm$2.04 \\
\cmidrule(lr){2-11}
 & W4 & AutoRound & 32.66$\pm$2.33 & 68.20$\pm$0.18 & 21.15$\pm$0.79 & 48.89$\pm$1.13 & 22.13$\pm$1.94 & 50.97$\pm$0.13 & 48.81$\pm$0.07 & 40.67$\pm$0.21 \\
 & W4 & \qselectta{} & 33.33$\pm$1.01 & 67.38$\pm$0.88 & 22.97$\pm$1.32 & 50.36$\pm$1.16 & 22.33$\pm$2.25 & 50.76$\pm$0.19 & 48.83$\pm$0.12 & 41.19$\pm$0.20 \\
\midrule
Llama-3.1-8B & BF16 & Reference & 28.28 & 79.01 & 25.25 & 60.45 & 16.40 & 63.23 & 59.68 & 45.44 \\
\cmidrule(lr){2-11}
 & W3 & AutoRound & 27.27$\pm$2.81 & 57.20$\pm$1.95 & 4.65$\pm$1.28 & 37.11$\pm$0.93 & 13.40$\pm$2.23 & 47.82$\pm$1.05 & 43.54$\pm$0.89 & 31.24$\pm$0.76 \\
 & W3 & \qselectta{} & 30.30$\pm$1.34 & 63.45$\pm$1.07 & 5.41$\pm$0.77 & 50.55$\pm$1.30 & 16.07$\pm$2.55 & 51.66$\pm$0.54 & 47.81$\pm$0.53 & 36.24$\pm$0.70 \\
\cmidrule(lr){2-11}
 & W4 & AutoRound & 31.82$\pm$1.01 & 73.89$\pm$0.43 & 27.37$\pm$2.33 & 55.44$\pm$0.54 & 16.07$\pm$1.70 & 59.49$\pm$0.20 & 56.38$\pm$0.27 & 44.01$\pm$0.58 \\
 & W4 & \qselectta{} & 27.95$\pm$2.87 & 76.48$\pm$0.94 & 24.18$\pm$4.59 & 59.02$\pm$1.93 & 14.07$\pm$1.51 & 60.69$\pm$0.31 & 57.34$\pm$0.14 & 43.73$\pm$0.23 \\
\midrule
Llama-3.2-3B & BF16 & Reference & 25.76 & 67.32 & 8.79 & 58.30 & 13.40 & 54.81 & 51.03 & 38.06 \\
\cmidrule(lr){2-11}
 & W3 & AWQ & 26.26$\pm$3.54 & 27.73$\pm$1.88 & 1.34$\pm$0.16 & 26.95$\pm$0.83 & 8.07$\pm$1.86 & 28.51$\pm$0.70 & 25.82$\pm$0.58 & 19.81$\pm$0.26 \\
 & W3 & GPTQ & 30.64$\pm$1.54 & 41.64$\pm$1.74 & 3.26$\pm$0.46 & 33.46$\pm$1.95 & 12.07$\pm$0.99 & 36.74$\pm$1.32 & 33.71$\pm$1.23 & 26.30$\pm$0.98 \\
 & W3 & AutoRound & 25.59$\pm$2.10 & 46.22$\pm$2.17 & 2.35$\pm$0.08 & 28.78$\pm$1.04 & 12.13$\pm$0.92 & 39.99$\pm$0.95 & 36.13$\pm$0.89 & 25.84$\pm$0.88 \\
 & W3 & \qselectta{} & 26.94$\pm$1.91 & 51.51$\pm$1.02 & 3.51$\pm$0.42 & 41.89$\pm$0.96 & 12.47$\pm$1.50 & 44.27$\pm$0.57 & 40.60$\pm$0.36 & 30.10$\pm$0.67 \\
 & W3 & \qselectkl{} & 26.09$\pm$0.29 & 50.88$\pm$0.79 & 3.59$\pm$0.70 & 39.75$\pm$2.94 & 11.60$\pm$1.25 & 42.47$\pm$0.63 & 39.02$\pm$0.49 & 29.06$\pm$0.42 \\
\cmidrule(lr){2-11}
 & W4 & AWQ & 30.13$\pm$1.05 & 64.82$\pm$0.44 & 10.06$\pm$1.40 & 54.10$\pm$1.18 & 12.73$\pm$1.42 & 51.82$\pm$0.09 & 48.46$\pm$0.20 & 37.28$\pm$0.54 \\
 & W4 & GPTQ & 25.08$\pm$1.77 & 63.88$\pm$0.56 & 9.33$\pm$2.36 & 49.28$\pm$1.08 & 11.00$\pm$1.06 & 52.16$\pm$0.21 & 48.24$\pm$0.15 & 35.12$\pm$0.65 \\
 & W4 & AutoRound & 26.26$\pm$1.75 & 63.74$\pm$0.84 & 7.53$\pm$2.09 & 48.44$\pm$2.13 & 13.00$\pm$1.40 & 51.63$\pm$0.49 & 47.71$\pm$0.50 & 35.10$\pm$0.80 \\
 & W4 & \qselectta{} & 25.93$\pm$1.77 & 64.25$\pm$0.17 & 8.52$\pm$1.38 & 51.50$\pm$0.62 & 13.73$\pm$1.21 & 52.49$\pm$0.88 & 48.66$\pm$0.80 & 36.07$\pm$0.67 \\
 & W4 & \qselectkl{} & 27.27$\pm$1.34 & 64.79$\pm$0.61 & 8.49$\pm$0.93 & 52.08$\pm$2.83 & 12.93$\pm$0.12 & 52.20$\pm$0.68 & 48.50$\pm$0.69 & 36.30$\pm$0.76 \\
\bottomrule
\end{tabular}
\end{adjustbox}}

\vspace{10pt}
{\setlength{\tabcolsep}{2.5pt}
\caption{Matched layer-wise scoring comparison. Both selectors use the same replicate-specific candidate bank, sequential layer-wise scaffold, 64$\times$1024-token budget, and three calibration-bank replicates. Values are mean $\pm$ sample standard deviation.}
\label{tab:matched_scoring}
\begin{adjustbox}{max width=\textwidth}
\begin{tabular}{llrrrrrrr}
\toprule
Setting & Score & Avg. PPL$\downarrow$ & Max-corpus PPL$\downarrow$ & Avg. KL$\downarrow$ & Max-corpus KL$\downarrow$ & Micro$\uparrow$ & Macro$\uparrow$ & Sel. min \\
\midrule
DS W3 & CALM-reimpl. (CKA) & 33.34$\pm$1.81 & 55.70$\pm$3.97 & 0.474$\pm$0.015 & 0.538$\pm$0.028 & 38.49$\pm$1.05 & 31.91$\pm$0.75 & 17.2 \\
 & Final-distribution KL & 32.87$\pm$0.80 & 55.12$\pm$2.04 & 0.427$\pm$0.006 & 0.476$\pm$0.013 & 39.23$\pm$1.24 & 33.08$\pm$1.38 & 21.6 \\
DS W4 & CALM-reimpl. (CKA) & 27.41$\pm$0.23 & 45.40$\pm$0.54 & 0.099$\pm$0.001 & 0.113$\pm$0.004 & 48.55$\pm$0.41 & 41.07$\pm$0.73 & 16.9 \\
 & Final-distribution KL & 27.97$\pm$0.17 & 46.67$\pm$0.53 & 0.094$\pm$0.001 & 0.107$\pm$0.002 & 48.52$\pm$0.30 & 40.61$\pm$0.48 & 20.4 \\
Llama-3.1 W3 & CALM-reimpl. (CKA) & 10.56$\pm$0.01 & 14.95$\pm$0.15 & 0.451$\pm$0.008 & 0.569$\pm$0.010 & 39.28$\pm$3.25 & 28.84$\pm$2.18 & 18.1 \\
 & Final-distribution KL & 9.89$\pm$0.11 & 14.07$\pm$0.26 & 0.365$\pm$0.002 & 0.472$\pm$0.019 & 44.73$\pm$1.70 & 33.13$\pm$1.25 & 19.4 \\
Llama-3.1 W4 & CALM-reimpl. (CKA) & 7.60$\pm$0.09 & 10.94$\pm$0.14 & 0.099$\pm$0.007 & 0.138$\pm$0.011 & 56.30$\pm$0.28 & 42.91$\pm$0.47 & 16.9 \\
 & Final-distribution KL & 7.54$\pm$0.03 & 10.81$\pm$0.03 & 0.082$\pm$0.001 & 0.113$\pm$0.001 & 56.98$\pm$0.51 & 43.79$\pm$0.45 & 21.2 \\
\bottomrule
\end{tabular}
\end{adjustbox}}

\vspace{10pt}
\begin{minipage}[t]{0.36\textwidth}\centering
{\caption{Paired Avg.-KL differences for the matched layer-wise scoring control. Values are final-distribution KL minus CALM-reimpl.; negative values favor final-distribution KL.}
\label{tab:matched_seed_deltas}
\begin{adjustbox}{max width=\linewidth}
\begin{tabular}{lrrr}
\toprule
Setting & Rep. 17 & Rep. 23 & Rep. 41 \\
\midrule
DS W3 & -0.0353 & -0.0543 & -0.0522 \\
DS W4 & -0.0058 & -0.0029 & -0.0062 \\
Llama-3.1 W3 & -0.0900 & -0.0788 & -0.0900 \\
Llama-3.1 W4 & -0.0158 & -0.0231 & -0.0118 \\
\bottomrule
\end{tabular}
\end{adjustbox}}
\end{minipage}\hfill
\begin{minipage}[t]{0.6\textwidth}\centering
{\caption{Objective variants relative to the default \qselectta{} aggregate mean. PPL and KL are computed as $100(\bar z_{\mathrm{variant}}-\bar z_{\mathrm{TA}})/\bar z_{\mathrm{TA}}$ from the three-replicate means; accuracy values are differences in percentage points. Negative PPL/KL and positive accuracy favor the variant.}
\label{tab:objective_ablation}
\begin{adjustbox}{max width=\linewidth}
\begin{tabular}{llrrrr}
\toprule
Setting & Configuration & $\Delta$PPL & $\Delta$KL & $\Delta$Micro & $\Delta$Macro \\
\midrule
DS W3 & \qselectkl{} & -4.8\% & -7.5\% & -0.15 & +0.75 \\
 & NLL-only & +18.6\% & +27.3\% & +1.23 & +1.30 \\
 & NLL-first + KL guard & +8.0\% & +9.0\% & +0.32 & +0.10 \\
DS W4 & \qselectkl{} & -1.7\% & -13.8\% & +0.02 & -0.45 \\
 & NLL-only & +6.8\% & +22.9\% & +0.23 & -0.18 \\
 & NLL-first + KL guard & +6.0\% & +14.2\% & -0.05 & -0.61 \\
Llama-3.1 W3 & \qselectkl{} & -2.1\% & -4.3\% & -0.90 & -1.20 \\
 & NLL-only & +3.6\% & +10.9\% & -1.20 & -1.30 \\
 & NLL-first + KL guard & +3.9\% & +10.7\% & -1.00 & -0.80 \\
Llama-3.1 W4 & \qselectkl{} & -0.6\% & -3.0\% & -0.30 & -0.40 \\
 & NLL-only & +1.8\% & +20.2\% & -0.80 & +0.30 \\
 & NLL-first + KL guard & +1.8\% & +20.2\% & -0.80 & +0.30 \\
Llama-3.2 W3 & \qselectkl{} & -5.5\% & -10.1\% & -1.57 & -1.03 \\
Llama-3.2 W4 & \qselectkl{} & -1.0\% & -4.5\% & -0.16 & +0.23 \\
\bottomrule
\end{tabular}
\end{adjustbox}}
\end{minipage}
\end{table*}

\begin{table*}[p]
\centering
\scriptsize
{\caption{Paired DS-R1-Qwen-7B W3 differences under replicate-specific candidate banks. Family counts are AWQ/AutoRound/GPTQ module assignments out of 196.}
\label{tab:dsw3_paired}
\begin{adjustbox}{max width=\textwidth}
\begin{tabular}{lrrrrrrrr}
\toprule
Rep. & Best-baseline PPL & $\Delta$PPL & $\Delta$PPL (\%) & Best-baseline KL & $\Delta$KL & $\Delta$Micro & $\Delta$Macro & A/AR/G \\
\midrule
17 & 32.9196 (GPTQ) & -2.0953 & -6.36 & 0.4307 (GPTQ) & -0.0072 & +1.11 & +2.11 & 12/146/38 \\
23 & 32.2026 (GPTQ) & -1.3345 & -4.14 & 0.4349 (GPTQ) & -0.0193 & +1.12 & +1.29 & 13/148/35 \\
41 & 34.2321 (AutoRound) & -3.8313 & -11.19 & 0.4410 (GPTQ) & -0.0197 & -2.47 & -1.10 & 13/146/37 \\
\bottomrule
\end{tabular}
\end{adjustbox}}

\vspace{10pt}
\begin{minipage}[t]{0.46\textwidth}\centering
{\caption{Search-budget and proxy controls relative to the main result. PPL/KL are relative percentages; accuracy values are percentage points.}
\label{tab:search_controls}
\begin{adjustbox}{max width=\linewidth}
\begin{tabular}{llrrrrr}
\toprule
Setting & Configuration & $\Delta$PPL & $\Delta$KL & $\Delta$Micro & $\Delta$Macro & Sel. min \\
\midrule
DS W3 & Three-pass & -0.75\% & -0.98\% & +0.62 & +0.75 & 443.6 \\
DS W4 & Three-pass & -0.12\% & -2.17\% & +0.09 & -0.18 & 432.1 \\
Llama-3.1 W3 & Three-pass & -0.32\% & -0.55\% & +0.34 & +0.52 & 567.1 \\
Llama-3.1 W4 & Three-pass & -0.13\% & -0.60\% & +0.25 & +0.49 & 545.8 \\
Llama-3.1 W3 & Shared-core proxy & -0.10\% & -0.57\% & -0.25 & -0.51 & 138.5 \\
Llama-3.1 W4 & Shared-core proxy & +0.04\% & +0.60\% & -0.29 & -0.37 & 155.6 \\
\bottomrule
\end{tabular}
\end{adjustbox}}
\end{minipage}\hfill
\begin{minipage}[t]{0.5\textwidth}\centering
{\caption{Initialization sensitivity under the default module-wise one-pass rule. Values are mean $\pm$ sample standard deviation over three calibration-bank replicates.}
\label{tab:initialization}
\begin{adjustbox}{max width=\linewidth}
\begin{tabular}{lllrrrr}
\toprule
Setting & Start & Avg. PPL$\downarrow$ & Avg. KL$\downarrow$ & Micro$\uparrow$ & Macro$\uparrow$ \\
\midrule
DS W3 & AutoRound & 30.70$\pm$0.26 & 0.420$\pm$0.004 & 41.77$\pm$1.95 & 34.02$\pm$2.04 \\
 & AWQ & 33.58$\pm$1.35 & 0.515$\pm$0.017 & 41.70$\pm$0.68 & 34.48$\pm$0.61 \\
 & GPTQ & 31.38$\pm$0.32 & 0.409$\pm$0.002 & 41.49$\pm$0.52 & 34.61$\pm$0.80 \\
DS W4 & AutoRound & 27.33$\pm$0.50 & 0.106$\pm$0.002 & 48.83$\pm$0.12 & 41.19$\pm$0.20 \\
 & AWQ & 27.18$\pm$0.23 & 0.113$\pm$0.004 & 48.99$\pm$0.18 & 41.13$\pm$0.43 \\
 & GPTQ & 27.69$\pm$0.13 & 0.092$\pm$0.002 & 48.87$\pm$0.14 & 40.81$\pm$0.90 \\
Llama-3.1 W3 & AutoRound & 9.79$\pm$0.05 & 0.349$\pm$0.008 & 47.81$\pm$0.53 & 36.24$\pm$0.70 \\
 & AWQ & 11.02$\pm$0.27 & 0.475$\pm$0.014 & 42.93$\pm$1.10 & 32.66$\pm$1.58 \\
 & GPTQ & 9.80$\pm$0.05 & 0.371$\pm$0.009 & 47.16$\pm$0.98 & 35.36$\pm$0.62 \\
Llama-3.1 W4 & AutoRound & 7.60$\pm$0.02 & 0.084$\pm$0.001 & 57.34$\pm$0.14 & 43.73$\pm$0.23 \\
 & AWQ & 7.61$\pm$0.02 & 0.096$\pm$0.001 & 57.50$\pm$0.35 & 44.81$\pm$1.04 \\
 & GPTQ & 7.54$\pm$0.03 & 0.087$\pm$0.001 & 56.70$\pm$0.48 & 42.82$\pm$0.95 \\
\bottomrule
\end{tabular}
\end{adjustbox}}
\end{minipage}

\vspace{10pt}
{\setlength{\tabcolsep}{2.7pt}
\caption{AWQ/GPTQ restricted-pool selector under two initializations. Values are mean $\pm$ sample standard deviation over the same three calibration-bank replicates as Table~\ref{tab:main_results}. Together with the full-pool AWQ/GPTQ rows in Table~\ref{tab:initialization}, these results form the pool-by-initialization control.}
\label{tab:pool_init}
\begin{adjustbox}{max width=\textwidth}
\begin{tabular}{llrrrrrr}
\toprule
Setting & Start & Avg. PPL$\downarrow$ & Max-corpus PPL$\downarrow$ & Avg. KL$\downarrow$ & Max-corpus KL$\downarrow$ & Micro$\uparrow$ & Macro$\uparrow$ \\
\midrule
DS W3 & AWQ & 34.83$\pm$0.99 & 58.53$\pm$2.20 & 0.563$\pm$0.012 & 0.654$\pm$0.013 & 39.33$\pm$1.50 & 33.93$\pm$0.67 \\
DS W3 & GPTQ & 31.70$\pm$0.63 & 53.16$\pm$1.67 & 0.432$\pm$0.002 & 0.485$\pm$0.007 & 38.87$\pm$0.23 & 32.85$\pm$1.30 \\
DS W4 & AWQ & 27.43$\pm$0.36 & 45.78$\pm$0.75 & 0.111$\pm$0.003 & 0.131$\pm$0.006 & 48.77$\pm$0.17 & 40.74$\pm$0.24 \\
DS W4 & GPTQ & 27.72$\pm$0.21 & 46.29$\pm$0.46 & 0.0928$\pm$0.0010 & 0.1059$\pm$0.0015 & 48.63$\pm$0.18 & 40.75$\pm$0.71 \\
Llama-3.1 W3 & AWQ & 11.40$\pm$0.23 & 16.25$\pm$0.56 & 0.519$\pm$0.012 & 0.669$\pm$0.024 & 38.91$\pm$1.43 & 29.12$\pm$1.73 \\
Llama-3.1 W3 & GPTQ & 10.22$\pm$0.08 & 14.22$\pm$0.14 & 0.419$\pm$0.007 & 0.590$\pm$0.009 & 44.39$\pm$0.43 & 34.26$\pm$0.90 \\
Llama-3.1 W4 & AWQ & 7.621$\pm$0.018 & 11.017$\pm$0.033 & 0.0983$\pm$0.0024 & 0.1282$\pm$0.0028 & 57.21$\pm$0.19 & 44.67$\pm$1.05 \\
Llama-3.1 W4 & GPTQ & 7.540$\pm$0.007 & 10.844$\pm$0.028 & 0.0882$\pm$0.0004 & 0.1299$\pm$0.0015 & 56.72$\pm$0.47 & 43.68$\pm$0.64 \\
\bottomrule
\end{tabular}
\end{adjustbox}}

\vspace{10pt}
{\setlength{\tabcolsep}{2.7pt}
\caption{Traversal-order sensitivity relative to the forward module-wise one-pass protocol. Values are paired mean $\pm$ sample standard deviation across calibration-bank replicates 17, 23, and 41.}
\label{tab:order_sensitivity}
\begin{adjustbox}{max width=\textwidth}
\begin{tabular}{llrrrrrr}
\toprule
Setting & Order & $\Delta$Avg. PPL & $\Delta$Max PPL & $\Delta$Avg. KL & $\Delta$Max KL & $\Delta$Micro & $\Delta$Macro \\
\midrule
DS W3 & Fixed random & +0.768$\pm$0.259 & +1.444$\pm$0.950 & +0.0183$\pm$0.0034 & +0.0238$\pm$0.0040 & +0.66$\pm$2.12 & +0.36$\pm$1.80 \\
DS W3 & Full reverse & +0.526$\pm$0.327 & +0.993$\pm$1.139 & +0.0041$\pm$0.0058 & +0.0038$\pm$0.0026 & +0.13$\pm$1.67 & +0.67$\pm$1.10 \\
DS W4 & Fixed random & +0.095$\pm$0.366 & +0.265$\pm$0.681 & -0.0004$\pm$0.0017 & +0.0008$\pm$0.0026 & -0.00$\pm$0.22 & -0.07$\pm$0.41 \\
DS W4 & Full reverse & +0.055$\pm$0.372 & +0.122$\pm$0.708 & +0.0010$\pm$0.0027 & +0.0022$\pm$0.0030 & -0.02$\pm$0.31 & -0.25$\pm$0.31 \\
Llama-3.1 W3 & Fixed random & -0.017$\pm$0.024 & +0.027$\pm$0.018 & -0.0044$\pm$0.0049 & -0.0040$\pm$0.0094 & +0.01$\pm$0.69 & +0.36$\pm$0.79 \\
Llama-3.1 W3 & Full reverse & +0.058$\pm$0.163 & +0.025$\pm$0.186 & +0.0000$\pm$0.0106 & +0.0074$\pm$0.0163 & -0.54$\pm$1.85 & -0.33$\pm$1.63 \\
Llama-3.1 W4 & Fixed random & -0.017$\pm$0.038 & -0.012$\pm$0.041 & -0.0000$\pm$0.0005 & -0.0006$\pm$0.0022 & -0.18$\pm$0.53 & -0.26$\pm$1.46 \\
Llama-3.1 W4 & Full reverse & -0.005$\pm$0.020 & -0.020$\pm$0.023 & -0.0009$\pm$0.0009 & -0.0011$\pm$0.0014 & -0.07$\pm$0.22 & -0.56$\pm$0.40 \\
\bottomrule
\end{tabular}
\end{adjustbox}}
\end{table*}


\begin{thebibliography}{99}
\raggedright
\bibitem{apiary2025} S.~Chapin, K.~Stewart, R.~Leontie, and C.~G. Henshaw, ``Autonomous Planning In-space Assembly Reinforcement-learning free-flYer (APIARY) International Space Station Astrobee testing,'' in \emph{Proc. Int. Conf. Space Robot. (iSpaRo)}, Sendai, Japan, 2025, pp.~418--424, doi:~10.1109/iSpaRo66239.2025.11436341.

\bibitem{ahn2022} B.~Ichter \emph{et al.}, ``Do as I can, not as I say: Grounding language in robotic affordances,'' in \emph{Proc. 6th Conf. Robot Learning}, \emph{Proc. Mach. Learn. Res.}, vol.~205, pp.~287--318, 2023. [Online]. Available: \url{https://proceedings.mlr.press/v205/ichter23a.html}

\bibitem{rt22023} B.~Zitkovich \emph{et al.}, ``RT-2: Vision-language-action models transfer web knowledge to robotic control,'' in \emph{Proc. 7th Conf. Robot Learning}, \emph{Proc. Mach. Learn. Res.}, vol.~229, pp.~2165--2183, 2023. [Online]. Available: \url{https://proceedings.mlr.press/v229/zitkovich23a.html}

\bibitem{openvla2024} M.~J.~Kim \emph{et al.}, ``OpenVLA: An open-source vision-language-action model,'' in \emph{Proc. 8th Conf. Robot Learning}, \emph{Proc. Mach. Learn. Res.}, vol.~270, pp.~2679--2713, 2025. [Online]. Available: \url{https://proceedings.mlr.press/v270/kim25c.html}

\bibitem{spacemind2026} A.~Wu, H.~Han, X.~Luo, R.~Wang, S.~He, and X.~Wan, ``SpaceMind: A modular and self-evolving embodied vision-language agent framework for autonomous on-orbit servicing,'' arXiv:2604.14399, 2026. [Online]. Available: \url{https://arxiv.org/abs/2604.14399}

\bibitem{furano2020} G.~Furano \emph{et al.}, ``Towards the use of artificial intelligence on the edge in space systems: Challenges and opportunities,'' \emph{IEEE Aerosp. Electron. Syst. Mag.}, vol.~35, no.~12, pp.~44--56, Dec. 2020, doi:~10.1109/MAES.2020.3008468.

\bibitem{nasa_perseverance_hw} NASA Science, ``Perseverance rover components,'' mission documentation. Accessed: Oct.~20, 2026. [Online]. Available: \url{https://science.nasa.gov/mission/mars-2020-perseverance/rover-components/}

\bibitem{nash2026radmit} J.~Nash, V.~Verma, E.~Graser, H.~Wang, G.~Kubiak, M.~Maimone, S.~Myint, J.~Biesiadecki, and P.~Romano, ``A software-based approach to radiation mitigation for planetary missions,'' in \emph{Proc. IEEE Aerosp. Conf.}, Big Sky, MT, USA, 2026, pp.~1--18, doi:~10.1109/AERO66936.2026.11519792.

\bibitem{wukong2025} Xinhua, ``Space station crew gains AI assistant,'' \emph{China Daily}, Aug.~18, 2025. Accessed: Oct.~20, 2026. [Online]. Available: \url{https://global.chinadaily.com.cn/a/202508/18/WS68a27b52a310b236346f217a.html}

\bibitem{mousist2025astrea} A.~D.~Mousist, ``ASTREA: Introducing agentic intelligence for orbital thermal autonomy,'' arXiv:2509.13380v2, Oct.~11, 2025. [Online]. Available: \url{https://arxiv.org/abs/2509.13380v2}

\bibitem{ecssml2024} European Cooperation for Space Standardization, ``Space engineering: Machine learning handbook,'' ECSS-E-HB-40-02A, Nov.~15, 2024. [Online]. Available: \url{https://ecss.nl/wp-content/uploads/2024/12/ECSS-E-HB-40-02A(15November2024).pdf}

\bibitem{frantar2023gptq} E.~Frantar, S.~Ashkboos, T.~Hoefler, and D.~Alistarh, ``GPTQ: Accurate post-training quantization for generative pre-trained transformers,'' in \emph{Proc. Int. Conf. Learn. Represent. (ICLR)}, 2023.

\bibitem{lin2024awq} J.~Lin, J.~Tang, H.~Tang, S.~Yang, W.-M. Chen, W.-C. Wang, G.~Xiao, X.~Dang, C.~Gan, and S.~Han, ``AWQ: Activation-aware weight quantization for LLM compression and acceleration,'' in \emph{Proc. Mach. Learn. Syst. (MLSys)}, 2024.

\bibitem{cheng2024signround} W.~Cheng, W.~Zhang, H.~Shen, Y.~Cai, X.~He, K.~Lv, and Y.~Liu, ``Optimize weight rounding via signed gradient descent for the quantization of LLMs,'' in \emph{Findings Assoc. Comput. Linguist.: EMNLP 2024}, 2024.

\bibitem{zhang2025calm} J.~Zhang, Y.~Zhang, and D.~Chen, ``CKA-guided modular quantization: Beyond bit-width to algorithmic diversity,'' arXiv:2512.16282, 2025.

\bibitem{kornblith2019cka} S.~Kornblith, M.~Norouzi, H.~Lee, and G.~Hinton, ``Similarity of neural network representations revisited,'' in \emph{Proc. Int. Conf. Mach. Learn. (ICML)}, 2019, pp.~3519--3529.

\bibitem{spacellava2025} M.~Foutter, D.~Gammelli, J.~Kruger, E.~Foss, P.~Bhoj, T.~Guffanti, S.~D'Amico, and M.~Pavone, ``Space-LLaVA: A vision-language model adapted to extraterrestrial applications,'' in \emph{Proc. IEEE Aerosp. Conf.}, Big Sky, MT, USA, 2025, pp.~1--23, doi:~10.1109/AERO63441.2025.11068757.

\bibitem{carrasco2025ksp} A.~Carrasco, V.~Rodriguez-Fernandez, and R.~Linares, ``Large language models as autonomous spacecraft operators in Kerbal Space Program,'' \emph{Adv. Space Res.}, vol.~76, no.~6, pp.~3480--3497, 2025, doi:~10.1016/j.asr.2025.06.034.

\bibitem{gr740} Frontgrade Gaisler, ``GR740: Quad-core LEON4FT microprocessor,'' product documentation. Accessed: Oct.~20, 2026. [Online]. Available: \url{https://www.gaisler.com/products/gr740}

\bibitem{nasa_hpsc2022} NASA, ``NASA awards next-generation spaceflight computing processor contract,'' Release 22-087, Aug.~15, 2022. Accessed: Oct.~20, 2026. [Online]. Available: \url{https://www.nasa.gov/news-release/nasa-awards-next-generation-spaceflight-computing-processor-contract/}

\bibitem{ciar2021} G.~Giuffrida \emph{et al.}, ``The $\Phi$-sat-1 mission: The first on-board deep neural network demonstrator for satellite Earth observation,'' \emph{IEEE Trans. Geosci. Remote Sens.}, vol.~60, Art.~no.~5517414, pp.~1--14, 2022, doi:~10.1109/TGRS.2021.3125567.

\bibitem{xqrversal} AMD, ``Radiation tolerant Versal AI Core series data sheet,'' DS946, ver.~1.2, Feb.~19, 2025. [Online]. Available: \url{https://docs.amd.com/r/en-US/ds946-xqr-versal-ai-core/General-Description}

\bibitem{thales_imagine2026} Thales Alenia Space, ``IMAGIN-e payload unlocks the potential of space edge computing aboard the ISS,'' Jan.~30, 2026. Accessed: Oct.~20, 2026. [Online]. Available: \url{https://www.thalesaleniaspace.com/en/news/imagin-e-payload-unlocks-potential-space-edge-computing-aboard-iss}

\bibitem{starcloud2025} P.~Singh, ```Greetings, earthlings': Nvidia-backed Starcloud trains first AI model in space as orbital data center race heats up,'' \emph{CNBC}, Dec.~10, 2025. Accessed: Oct.~20, 2026. [Online]. Available: \url{https://www.cnbc.com/2025/12/10/nvidia-backed-starcloud-trains-first-ai-model-in-space-orbital-data-centers.html}

\bibitem{opssat_esa} European Space Agency, ``OPS-SAT,'' mission documentation. Accessed: Oct.~20, 2026. [Online]. Available: \url{https://www.esa.int/Enabling_Support/Operations/OPS-SAT}

\bibitem{nagel2020adaround} M.~Nagel, R.~A. Amjad, M.~van Baalen, C.~Louizos, and T.~Blankevoort, ``Up or down? Adaptive rounding for post-training quantization,'' in \emph{Proc. Int. Conf. Mach. Learn. (ICML)}, 2020.

\bibitem{xiao2023smoothquant} G.~Xiao, J.~Lin, M.~Seznec, H.~Wu, J.~Demouth, and S.~Han, ``SmoothQuant: Accurate and efficient post-training quantization for large language models,'' in \emph{Proc. Int. Conf. Mach. Learn. (ICML)}, 2023.

\bibitem{shao2024omniquant} W.~Shao, M.~Chen, Z.~Zhang, P.~Xu, L.~Zhao, Z.~Li, K.~Zhang, P.~Gao, Y.~Qiao, and P.~Luo, ``OmniQuant: Omnidirectionally calibrated quantization for large language models,'' in \emph{Proc. Int. Conf. Learn. Represent. (ICLR)}, 2024.

\bibitem{ashkboos2024quarot} S.~Ashkboos, A.~Mohtashami, M.~L. Croci, B.~Li, P.~Cameron, M.~Jaggi, D.~Alistarh, T.~Hoefler, and J.~Hensman, ``QuaRot: Outlier-free 4-bit inference in rotated LLMs,'' in \emph{Adv. Neural Inf. Process. Syst. (NeurIPS)}, 2024.

\bibitem{gu2026flrq} H.~Gu, L.~Hu, S.~Niu, and F.~Liu, ``FLRQ: Faster LLM quantization with flexible low-rank matrix sketching,'' in \emph{Proc. AAAI Conf. Artif. Intell.}, vol.~40, no.~26, 2026, pp.~21369--21377.

\bibitem{zhao2026specquant} Z.~Zhao, F.~Liu, J.~Wang, C.~Guan, Z.~Wang, L.~Jiang, and H.~Guan, ``SpecQuant: Spectral decomposition and adaptive truncation for ultra-low-bit LLMs quantization,'' in \emph{Proc. AAAI Conf. Artif. Intell.}, vol.~40, no.~34, 2026, pp.~28786--28794.

\bibitem{akbulut2026infoq} M.~E. Akbulut, H.~H.~Y. Shalby, F.~Pittorino, and M.~Roveri, ``InfoQ: Mixed-precision quantization via global information flow,'' in \emph{Proc. AAAI Conf. Artif. Intell.}, vol.~40, no.~24, 2026, pp.~19598--19606.

\bibitem{wang2019haq} K.~Wang, Z.~Liu, Y.~Lin, J.~Lin, and S.~Han, ``HAQ: Hardware-aware automated quantization with mixed precision,'' in \emph{Proc. IEEE/CVF Conf. Comput. Vis. Pattern Recognit. (CVPR)}, 2019.

\bibitem{dong2019hawq} Z.~Dong, Z.~Yao, A.~Gholami, M.~W. Mahoney, and K.~Keutzer, ``HAWQ: Hessian aware quantization of neural networks with mixed-precision,'' in \emph{Proc. IEEE/CVF Int. Conf. Comput. Vis. (ICCV)}, 2019.

\bibitem{hinton2015distilling} G.~Hinton, O.~Vinyals, and J.~Dean, ``Distilling the knowledge in a neural network,'' arXiv:1503.02531, 2015.

\bibitem{edalati2025oac} A.~Edalati, A.~Ghaffari, M.~Ghazvini Nejad, L.~Hou, B.~Chen, M.~Asgharian, and V.~Partovi Nia, ``OAC: Output-adaptive calibration for accurate post-training quantization,'' in \emph{Proc. AAAI Conf. Artif. Intell.}, vol.~39, no.~16, 2025, pp.~16453--16461.

\bibitem{li2021brecq} Y.~Li, R.~Gong, X.~Tan, Y.~Yang, P.~Hu, Q.~Zhang, F.~Yu, W.~Wang, and S.~Gu, ``BRECQ: Pushing the limit of post-training quantization by block reconstruction,'' in \emph{Proc. Int. Conf. Learn. Represent. (ICLR)}, 2021.

\bibitem{bansal2021stitching} Y.~Bansal, P.~Nakkiran, and B.~Barak, ``Revisiting model stitching to compare neural representations,'' in \emph{Adv. Neural Inf. Process. Syst. (NeurIPS)}, vol.~34, 2021, pp.~225--236.

\bibitem{guo2025deepseekr1} D.~Guo \emph{et al.}, ``DeepSeek-R1: Incentivizing reasoning capability in LLMs via reinforcement learning,'' arXiv:2501.12948, 2025.

\bibitem{deepseek2025modelcard} DeepSeek-AI, ``DeepSeek-R1-Distill-Qwen-7B model card,'' 2025. [Online]. Available: \url{https://huggingface.co/deepseek-ai/DeepSeek-R1-Distill-Qwen-7B}

\bibitem{dubey2024llama3} A.~Dubey \emph{et al.}, ``The Llama 3 herd of models,'' arXiv:2407.21783, 2024.

\bibitem{meta2024llama32} Meta AI, ``Llama 3.2 model card,'' 2024. [Online]. Available: \url{https://github.com/meta-llama/llama-models/blob/main/models/llama3_2/MODEL_CARD.md}

\bibitem{raffel2020exploring} C.~Raffel, N.~Shazeer, A.~Roberts, K.~Lee, S.~Narang, M.~Matena, Y.~Zhou, W.~Li, and P.~J. Liu, ``Exploring the limits of transfer learning with a unified text-to-text transformer,'' \emph{J. Mach. Learn. Res.}, vol.~21, no.~140, pp.~1--67, 2020.

\bibitem{merity2016pointer} S.~Merity, C.~Xiong, J.~Bradbury, and R.~Socher, ``Pointer sentinel mixture models,'' arXiv:1609.07843, 2016.

\bibitem{rein2023gpqa} D.~Rein, B.~L. Hou, A.~C. Stickland, J.~Petty, R.~Y. Pang, J.~Dirani, J.~Michael, and S.~R. Bowman, ``GPQA: A graduate-level Google-proof Q\&A benchmark,'' arXiv:2311.12022, 2023.

\bibitem{clark2018arc} P.~Clark, I.~Cowhey, O.~Etzioni, T.~Khot, A.~Sabharwal, C.~Schoenick, and O.~Tafjord, ``Think you have solved question answering? Try ARC, the AI2 reasoning challenge,'' arXiv:1803.05457, 2018.

\bibitem{cobbe2021gsm8k} K.~Cobbe \emph{et al.}, ``Training verifiers to solve math word problems,'' arXiv:2110.14168, 2021.

\bibitem{zellers2019hellaswag} R.~Zellers, A.~Holtzman, Y.~Bisk, A.~Farhadi, and Y.~Choi, ``HellaSwag: Can a machine really finish your sentence?'' in \emph{Proc. 57th Annu. Meeting Assoc. Comput. Linguist. (ACL)}, 2019.

\bibitem{lightman2024verify} H.~Lightman, V.~Kosaraju, Y.~Burda, H.~Edwards, B.~Baker, T.~Lee, J.~Leike, J.~Schulman, I.~Sutskever, and K.~Cobbe, ``Let's verify step by step,'' in \emph{Proc. Int. Conf. Learn. Represent. (ICLR)}, 2024.

\bibitem{hendrycks2021mmlu} D.~Hendrycks, C.~Burns, S.~Basart, A.~Zou, M.~Mazeika, D.~Song, and J.~Steinhardt, ``Measuring massive multitask language understanding,'' in \emph{Proc. Int. Conf. Learn. Represent. (ICLR)}, 2021.

\bibitem{wang2024mmlupro} Y.~Wang \emph{et al.}, ``MMLU-Pro: A more robust and challenging multi-task language understanding benchmark,'' arXiv:2406.01574, 2024.
\end{thebibliography}
\end{document}